\documentclass[journal]{IEEEtran}

\usepackage{amsmath,amssymb}
\usepackage{graphicx}
\usepackage{algorithm}
\usepackage{algpseudocode}
\usepackage{booktabs}
\usepackage{multirow}
\usepackage[utf8]{inputenc}
\usepackage[T1]{fontenc}
\usepackage{textcomp}
\usepackage{xcolor}
\usepackage{url}
\usepackage{cite}
\usepackage{pifont}
\usepackage{tikz}
\usetikzlibrary{arrows.meta,positioning}

\graphicspath{{figures/}}

\begin{document}

\title{Training-Free Human-in-the-Loop Anomaly Detection via Memory Bank Correction}

\author{Ayusha~Abbas,~Saram~Abbas,~and~Kabita~Adhikari
\thanks{A. Abbas, S. Abbas, and K. Adhikari are with the School of Engineering, Newcastle University, Newcastle upon Tyne, UK (e-mail: ayusha.abbas24@gmail.com; s.abbas11@newcastle.ac.uk; kabita.adhikari@newcastle.ac.uk).}}

\markboth{Preprint}{Abbas \MakeLowercase{\textit{et al.}}: Training-Free Human-in-the-Loop Anomaly Detection via Memory Bank Correction}

\maketitle

\begin{abstract}
Anomaly detectors are hardest to deploy exactly where training data is scarcest: a newly commissioned production line has a handful of verified ``golden'' samples and no machine-learning engineer on the factory floor. We present a training-free human-in-the-loop framework in which a domain expert corrects a PatchCore detector by direct memory bank editing: no retraining, no gradients, no original training data. A false-positive correction inserts the reviewed image's normal patches through a self-calibrating \emph{novelty gate} admitting only those beyond the median pool-normal nearest-neighbour distance. From a bank built on only ten golden samples, operator corrections close a median 66\% of the gap to an uncorrected fully trained bank (mean 80\%, raised by three categories that overshoot parity), significantly improving 12 of 15 MVTec AD categories and harming none: ten samples plus corrections outperform hundreds of samples without them. On already-trained banks the headroom is smaller and concentrated where the bank undersamples normal appearance (gated: toothbrush $+0.10$, metal nut $+0.09$, zipper $+0.05$, screw $+0.05$), and no category except grid is significantly harmed. Evaluation uses a held-out protocol (20 splits per category, Holm-corrected Wilcoxon), because corrected images entering the bank inflate naive evaluation toward AUROC 1.0 by memorisation. Passive and active querying are statistically indistinguishable; a matched-label-budget control attributes gains to deployment-time label production at 43\% of exhaustive-review cost; a defect-memory extension fails decisively. Feedback is simulated from ground truth; live expert trials, where mislabelling is costliest on small banks, remain future work.
\end{abstract}


\section{Introduction}

Deep anomaly detection now achieves near-perfect discrimination on industrial inspection benchmarks \cite{roth2022patchcore,bergmann2019mvtec}, but deployed models do not stay accurate. Normal product appearance drifts (new component variants, changed illumination, camera recalibration) and errors accumulate. The standard remedy, periodic retraining, requires new labelled data, training infrastructure, and machine learning (ML) expertise that is rarely present on the factory floor. The people who \emph{can} retrain the model (ML engineers) are not the people who see its errors every day (quality engineers), so in regulated manufacturing environments a model correction can take weeks to months of escalation, retraining, and revalidation.

\begin{figure*}[!t]
\centering
\includegraphics[width=\textwidth]{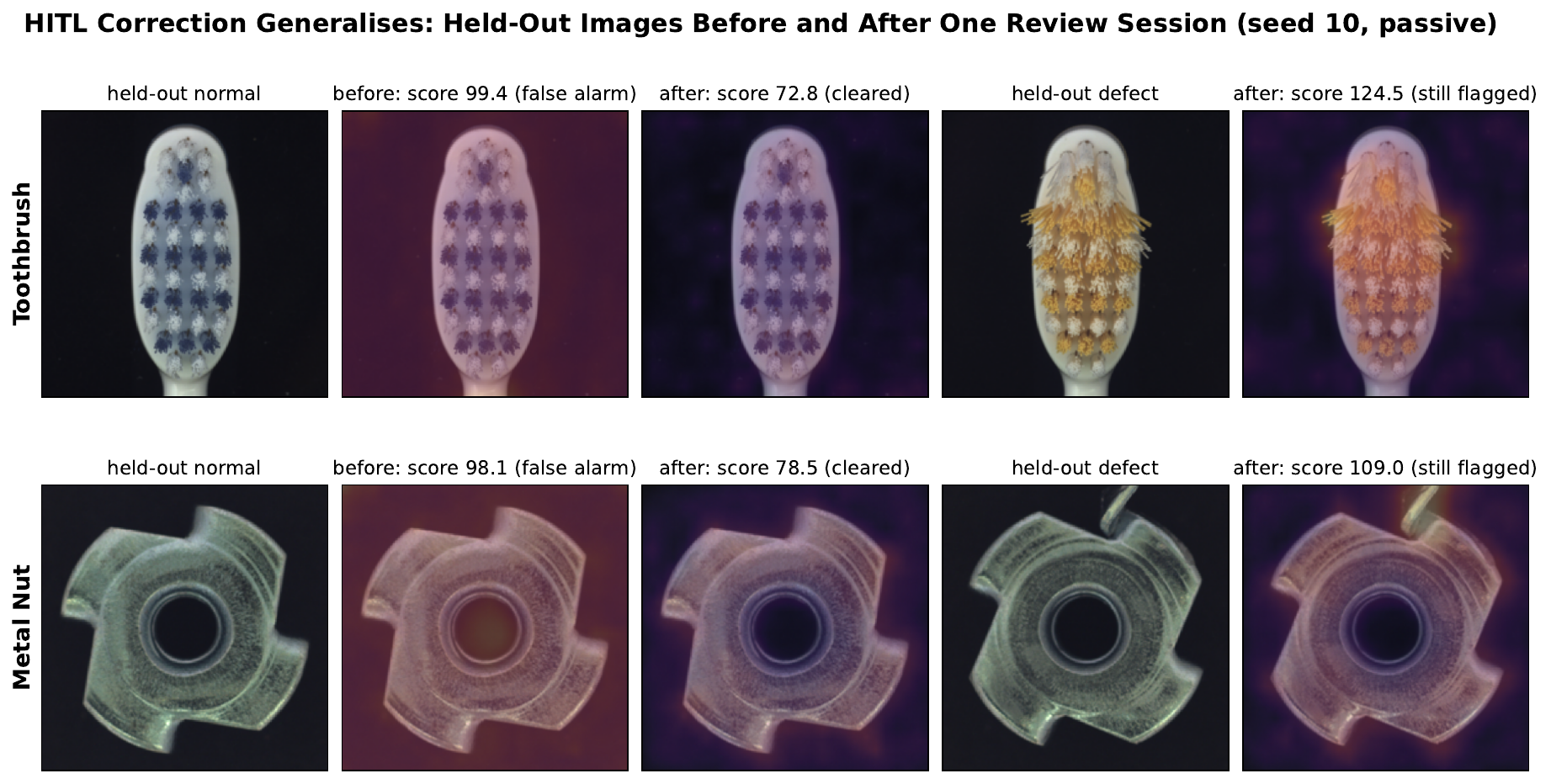}
\caption{What training-free correction looks like on images the loop never touches. For toothbrush and metal nut, a \emph{held-out} normal image is falsely flagged before correction (anomaly heatmap hot everywhere) and cleared after a single simulated review session (16--25 single-decision reviews for the seeds shown; the category means are $\sim$13 and $\sim$20), while a held-out defect remains confidently flagged, with the anomaly map concentrated on the defect itself (contaminated bristles; flipped part). Heatmaps share a colour scale within each row; scores are the image-level anomaly scores.}
\label{fig:qualitative}
\end{figure*}

A second, sharper version of the same problem is \textbf{cold start}: a newly commissioned production line has almost no training data at all. In both situations the quality engineer already knows what the model does not: which flagged images are genuine defects and which are false alarms. Existing systems give them no way to put that knowledge into the model.

Detection failures carry industrial-scale costs: automotive original equipment manufacturers (OEMs) paid USD~51 billion in warranty claims in 2023 \cite{warrantyweek2024}, a single end-of-line contamination excursion in semiconductor fabrication can exceed USD~21 million \cite{hillier2014expensive}, and a defective-medicine prosecution carries six-figure penalties \cite{mhra2022prosecution}. All three are downstream costs of defects that escaped inspection, and a defect grows more expensive at every stage it survives. A detector that degrades silently therefore does not just lose accuracy; it feeds escaped defects into the most expensive end of this cost structure.

Benchmark ceilings do not close this gap. PatchCore's headline 99.6\% area under the receiver operating characteristic (ROC) curve (AUROC) on MVTec AD requires an ensemble of three large backbones at high resolution \cite{roth2022patchcore}; the leaner configurations that fit factory edge hardware give up several points of accuracy (our single-backbone deployment configuration averages 93.4\%), and even the 99.6\% ensemble, classifying 50{,}000 images a day, hands the quality team 200 misclassifications every day. PatchCore's authors themselves list deployment-time adaptation as an open problem \cite{roth2022patchcore}.

This paper closes that gap with a \textbf{training-free human-in-the-loop (HITL) framework}: the quality engineer reviews a flagged error, makes the single binary judgment they already make daily (``normal'' or ``defect''), and the framework applies that judgment as a direct edit to the model's knowledge store, effective immediately. It needs no retraining, no ML engineer, and no access to the original training data. Figure~\ref{fig:qualitative} previews the outcome: one short review session clears a false alarm \emph{on images the loop never saw}, while genuine defects remain flagged.

PatchCore \cite{roth2022patchcore} is uniquely suited to this because its knowledge is not buried in network weights: it stores normal appearance as an explicit \emph{memory bank} of patch features. Adding patches from a falsely flagged normal image extends what the model accepts as normal; removing patches near a missed defect raises future scores for similar defects. Each correction is an interpretable edit to a data structure. Prior work modifies PatchCore banks automatically, via continual learning \cite{bugarin2024unveiling} or loss-driven optimisation \cite{jiang2024frpatchcore}, but still requires ML infrastructure; to our knowledge, ours is the first framework in which a domain expert corrects specific deployment-time misclassifications by direct memory bank surgery.

\subsection{Contributions}

We evaluate the framework on all 15 MVTec AD categories \cite{bergmann2019mvtec}, 20 independent splits each, under Holm-corrected Wilcoxon tests. Our contributions:

\begin{itemize}
\item \textbf{Cold-start recovery from ten samples.} From a bank built on only 10 golden samples, operator corrections close a median 66\% of the gap to an uncorrected fully trained bank (mean 80\%, raised by three categories that overshoot parity), significantly improving 12 of 15 categories and harming none. Ten samples plus corrections outperform hundreds of samples without them. We do not claim curation dominates data: a corrected full-training bank still finishes higher (toothbrush 0.927 vs.\ 0.904, metal nut 0.989 vs.\ 0.927). The claim is about what is achievable on day one of a deployment, and this is the one setting in which every category either improves or is left untouched.

\item \textbf{A training-free correction mechanism.} False positives (FPs) and false negatives (FNs) are resolved by direct memory bank edits, with no retraining, gradients, or training data. An FP/FN ablation shows FP insertion reproduces the entire gain while FN removal is a measured near no-op: the mechanism works by \emph{teaching normality}, not by memorising defects.

\item \textbf{A self-calibrating novelty gate.} A patch is inserted only if its nearest-neighbour distance to the bank exceeds the median pool-normal distance. On already-trained banks the gate eliminates the held-out degradation the raw loop causes on five of six well-covered categories, preserves every category that benefits, and is robust across a quantile sensitivity sweep, leaving grid as the single category still significantly harmed.

\item \textbf{A held-out evaluation protocol} (plus AUCC, the area-under-the-correction-curve efficiency metric): corrected images enter the bank, so naive evaluation on reviewed images inflates toward AUROC 1.0 by memorisation. Correcting a review pool while scoring a disjoint held-out set separates generalisation from memorisation, and reveals six categories where the ungated loop is flat-to-negative on mature banks, a boundary invisible to pool-only evaluation that we report as an explicit negative result. A matched-label-budget control shows the gains equal what the same labels would achieve at training time: the loop's value is producing those labels at deployment, at 43\% of the cost of exhaustive review.

\item \textbf{A strategy equivalence result.} Passive (random) and active (uncertainty) querying are statistically indistinguishable on every category, because bank edits act globally and the review budget exhausts the error pool under either ordering. No acquisition function is needed.
\end{itemize}

In summary: from ten golden samples a domain expert with no ML knowledge closes most of the gap to a fully trained detector, improving 12 of 15 categories and harming none, in tens of single-decision reviews. On mature banks the same loop raises held-out AUROC on the four categories whose banks undersample normal appearance, from 0.826 to 0.972 (toothbrush) and 0.900 to 0.998 (metal nut) in mean sessions of $\sim$13 and $\sim$20 reviews; those figures are for the ungated loop, and under the novelty-gated configuration we recommend for deployment the same categories reach 0.927 and 0.989, with no category except grid significantly harmed, so the loop is safe to leave on.

\section{Related Work}
\label{sec:related_work}

Our work sits at the intersection of unsupervised anomaly detection, human-in-the-loop machine learning, and active learning query strategies.

\subsection{Unsupervised Anomaly Detection}

Deep learning-based anomaly detection spans a broad taxonomy of reconstruction-based, density-based, and discriminative approaches \cite{pang2021deep}. Early anomaly detection approaches relied on reconstruction-based methods, such as convolutional autoencoders and generative adversarial networks (GANs) like AnoGAN \cite{schlegl2017anogan}, which detect anomalies by their high reconstruction error or latent space distance. These suffer from mode collapse, blurry reconstructions of high-frequency textures, and high computational overhead. A second family exploits pretrained convolutional neural network (CNN) features \cite{napoletano2018anomaly}, showing that ImageNet representations transfer to industrial inspection without domain-specific training. Student-teacher frameworks \cite{wang2021stfpm} extend this by detecting anomalies through student-teacher feature discrepancy.

The state of the art is dominated by memory bank methods. SPADE \cite{cohen2020spade} showed that nearest-neighbour comparison against a bank of pretrained features achieves strong localisation with no training; PatchCore \cite{roth2022patchcore} refines this with a greedy coreset of locally-aware patch features, reaching 99.6\% mean AUROC on MVTec AD with a three-backbone ensemble at 320$\times$320 resolution. That figure is a static benchmark result, and PatchCore's authors explicitly leave deployment-time adaptation as future work. Our framework builds on PatchCore's architecture and answers that gap. PaDiM \cite{defard2021padim} and SimpleNet \cite{liu2023simplenet} are related feature-based alternatives.

\subsection{Human-in-the-Loop Machine Learning}

Human-in-the-loop machine learning \cite{monarch2021hitl} incorporates human feedback into the learning process, classically via active learning \cite{settles2010active}. In visual inspection, Ro{\v{z}}anec et al.\ \cite{rozanec2023human} combine active learning with explainable artificial intelligence (XAI) heat maps to reduce annotation burden for supervised retraining; HILAD \cite{deng2024hilad} proposes bidirectional HITL for time-series anomaly detection. Within anomaly detection, Das et al.\ \cite{das2018active,das2024tree} and Bodor et al.\ \cite{bodor2022active} use expert feedback to re-weight or re-optimise detectors, so the feedback still flows through retraining. Closest to our setting, Vishwakarma et al.\ \cite{vishwakarma2024taming} use human feedback on flagged samples to adapt the \emph{decision threshold} with false-positive-rate guarantees, but never modify the representation; a follow-up additionally adapts a learned scoring network, but through periodic gradient optimisation \cite{yamada2025adaptive}. We differ on all three axes: no retraining at all, corrections that edit the representation of normality itself, and responsibility shifted from ML engineers to the quality engineers who already review the errors. To our knowledge, ours is the first framework enabling human-driven, training-free correction of specific misclassified samples through direct memory bank surgery at deployment time.

\subsection{Active Learning and Query Strategies}

Uncertainty sampling \cite{lewis1994sequential}, the standard heuristic in active learning, selects the sample about which the model is least confident. Query-by-committee \cite{seung1992query} and density-weighted methods \cite{settles2010active} provide alternative strategies. A consistent finding is that uncertainty-based strategies outperform passive random sampling \cite{settles2010active}. Our work tests this regularity for memory bank correction and finds that it does not transfer: under multi-seed held-out evaluation, uncertainty sampling fails to outperform random sampling, and we provide a geometric explanation for why. This finding has implications for both active learning theory and the design of HITL anomaly detection systems.

\subsection{Continual Learning and Test-Time Adaptation}

Continual learning \cite{delange2022continual} and test-time adaptation \cite{wang2021tent} address deployment-time improvement through parameter updates, requiring careful regularisation against catastrophic forgetting. Within the PatchCore literature specifically, continual learning variants have emerged: \textit{PatchCoreCL} \cite{bugarin2024unveiling} maintains per-task memory banks updated as new tasks arrive, and \textit{FR-PatchCore} \cite{jiang2024frpatchcore} updates the memory bank during training via a negative cosine similarity loss. A growing line of continual industrial anomaly detection likewise treats the memory bank or its coreset as the unit of adaptation: Li et al.\ \cite{li2022continual} adapt the detector as new normal samples arrive under distribution shift, CADIC \cite{yang2025cadic} incrementally updates embeddings within a fixed-size unified coreset shared across tasks, MECAD \cite{dahmardeh2025mecad} assigns per-class experts with coreset replay buffers, and orthogonal LoRA banks \cite{fang2026lora} isolate per-task adaptation in low-rank parameter subspaces to preserve learned normality. These approaches share our insight that the memory bank is the modifiable component of memory-based detectors, but differ fundamentally in the update signal: all are driven by a stream of new, assumed-clean training data arriving at task or batch boundaries, whereas our framework is driven by an operator's dispositions of \emph{individual misclassified samples}: error-targeted corrections at deployment time with no new training set, no optimisation objective, and no assumption about task structure. Mechanistically closest to our novelty gate (Section~\ref{sec:gating}), TimeRep \cite{han2025timerep} augments a time-series memory bank at inference with representations whose nearest-neighbour distance exceeds a quantile threshold, but as unsupervised drift tracking over a stream, not as gating of human-verified corrections.

Few-shot anomaly detection builds the detector itself from $k$ samples: RegAD meta-learns a registration proxy across categories \cite{huang2022regad}, WinCLIP transfers language-guided features zero-/few-shot \cite{jeong2023winclip}, and GraphCore \cite{xie2023graphcore} and FastRecon \cite{fang2023fastrecon} build compact memory banks directly from the $k$ normals. All treat the residual gap to full-training performance as fixed once deployed. Our cold-start evaluation (Section~\ref{sec:coldstart_protocol}) starts in the same regime, with a bank built from ten samples, and measures how much of that gap operator corrections then close; the contribution is the recovery-by-correction protocol, not a new few-shot detector. The framework itself is orthogonal: it applies on top of any memory bank model as a deployment-time correction layer.

\subsection{Positioning}

Table~\ref{tab:positioning} summarises how our approach relates to existing paradigms across four key desiderata.

\begin{table}[!t]
\caption{Positioning of our work relative to prior methods.}
\label{tab:positioning}
\centering
\begin{tabular}{lcccc}
\toprule
\textbf{Method} & \textbf{No re-} & \textbf{Human} & \textbf{Deploy-} & \textbf{Specific} \\
 & \textbf{training} & \textbf{feedback} & \textbf{ment-time} & \textbf{correction} \\
\midrule
Autoencoder         & \ding{55} & \ding{55} & \ding{55} & \ding{55} \\
AnoGAN               & \ding{55} & \ding{55} & \ding{55} & \ding{55} \\
PatchCore            & \ding{51} & \ding{55} & \ding{55} & \ding{55} \\
Few-shot AD          & Partial & \ding{55} & Partial & \ding{55} \\
Continual learning   & \ding{55} & \ding{55} & \ding{51} & \ding{55} \\
Test-time adapt.     & Partial & \ding{55} & \ding{51} & \ding{55} \\
HITL + retraining    & \ding{55} & \ding{51} & \ding{55} & Partial \\
\textbf{Ours}        & \ding{51} & \ding{51} & \ding{51} & \ding{51} \\
\bottomrule
\end{tabular}
\end{table}

\section{Methodology}
\label{sec:methodology}

We first summarise the PatchCore baseline, then present the HITL correction mechanism (including the novelty gate), the querying strategies, the AUCC metric, and the experimental protocols.

\subsection{PatchCore Baseline}

\subsubsection{Features and Memory Bank}

PatchCore \cite{roth2022patchcore} represents normal appearance as an explicit memory bank of locally-aware patch features from a frozen WideResNet-50 \cite{zagoruyko2016wide} pretrained on ImageNet \cite{deng2009imagenet}. Feature maps are taken from backbone levels 2 (28$\times$28, 512 channels; fine texture) and 3 (14$\times$14, 1{,}024 channels; part-level structure): level 3 is bilinearly upsampled to 28$\times$28 and concatenated channel-wise with level 2, so each of the $28 \times 28 = 784$ spatial positions carries a 1{,}536-dimensional descriptor $p_{ij}(x)$. We write $P(x)$ for that set, $|P(x)| = 784$. Backbone weights are fixed throughout coreset construction and every HITL experiment; no gradient is computed at any point.

The union of patch features over a category's defect-free training images is too large to query directly (some 200{,}000 vectors for a typical 200--300 image category), so PatchCore compresses it by greedy $k$-center coreset subsampling \cite{sener2018coreset}, retaining the subset that minimises the maximum distance from any training patch to its nearest retained entry. We keep 1\% of training patches, giving banks $M \in \mathbb{R}^{N \times 1536}$ ranging from $N = 470$ (toothbrush, 60 training images) to 3{,}065 (hazelnut, 391). How undersampled a category's bank is turns out to predict where correction helps (Section~\ref{sec:results}).

\subsubsection{Anomaly Scoring}

At inference time, for a test image $x$, we extract its 784 patch features $P(x)$ and compute the anomaly score as the maximum patch-level distance to the memory bank, each patch's distance being the mean over its $k=9$ nearest bank entries:

\begin{equation}
s(x) = \max_{p \in P(x)} d_9(p), \qquad d_9(p) = \frac{1}{9} \sum_{m \in \mathcal{N}_9(p, M)} \lVert p - m \rVert_2
\label{eq:anomaly_score}
\end{equation}

where $\mathcal{N}_9(p, M)$ denotes the nine nearest entries of $M$ to $p$. The max identifies the single most anomalous patch, making the score sensitive to the small local defects typical of industrial inspection, and the same nine distances form the anomaly map used for localisation. Averaging over a small neighbourhood, rather than taking the 1-NN distance, makes the per-patch value less sensitive to a single unrepresentative bank entry. Retrieval uses a ball-tree index.

An image is classified anomalous if $s(x) \ge \tau$, with $\tau$ the 90th percentile of scores on normal training images (no anomalous examples needed), recomputed after every correction round. Our implementation is a single WideResNet-50 at 224$\times$224 without score reweighting, and departs from canonical PatchCore in using the mean-of-nine patch statistic of Eq.~\ref{eq:anomaly_score} in place of the 1-NN score; this is a deliberate implementation choice, applied identically to every baseline and every corrected bank reported here. It gives 0.934 mean AUROC, deliberately below PatchCore's 99.6\% ensemble headline, because a model at 99.6\% leaves no headroom for studying human-driven correction, whereas this deployable configuration has realistic error pools.

\subsection{Human-in-the-Loop Feedback Framework}

\begin{figure*}[!t]
\centering
\begin{tikzpicture}[
  font=\small,
  node distance=5mm and 8mm,
  box/.style={draw, rounded corners=2pt, align=center, minimum height=9mm, inner sep=5pt},
  accent/.style={box, thick, fill=blue!8},
  human/.style={box, thick, fill=orange!15},
  faded/.style={box, draw=gray!80, text=black!60, fill=gray!8, dashed},
  arr/.style={-{Stealth[length=2.6mm]}, thick},
  lbl/.style={font=\scriptsize, align=center, midway}
]
\node[box] (cam) {Production\\images};
\node[accent, right=of cam] (model) {\textbf{Deployed PatchCore}\\frozen backbone\\$+$ memory bank $M$};
\node[box, right=of model] (pred) {Prediction:\\normal / defect};
\node[human, right=of pred] (expert) {\textbf{Quality engineer}\\one binary decision:\\``normal'' or ``defect''};
\node[accent, below=8mm of pred] (corr) {\textbf{Direct memory bank edit}: no retraining, no gradients, no training data\\``normal'' (FP) $\to$ novelty-gated patch insertion into $M$ (Eq.~\ref{eq:gating}) \qquad ``defect'' (FN) $\to$ remove $k$ nearest bank entries};
\node[faded, below=6mm of corr] (retrain) {Conventional remedy: collect data $\to$ label $\to$ retrain $\to$ validate $\to$ redeploy \quad (ML team, retraining infrastructure, days--weeks)};
\draw[arr] (cam) -- (model);
\draw[arr] (model) -- (pred);
\draw[arr] (pred) -- node[lbl, above]{flagged\\errors} (expert);
\draw[arr] (expert.south) |- node[lbl, above right=4mm and 2mm]{one label} (corr.east);
\draw[arr] (corr.north -| model.south) -- node[lbl, right=1mm]{immediate effect} (model.south);
\end{tikzpicture}
\caption{Framework overview. A deployed PatchCore model flags errors; a quality engineer (not an ML engineer) resolves each with a single binary decision, which is applied as a direct edit to the memory bank and takes effect immediately. The conventional remedy (bottom, dashed) requires an ML team and a retraining pipeline.}
\label{fig:overview}
\end{figure*}
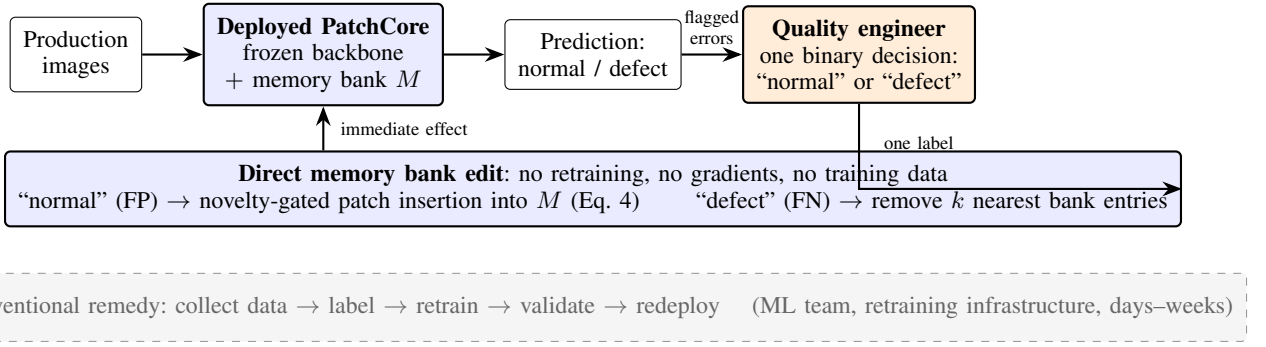

Our framework (Fig.~\ref{fig:overview}) is a post-deployment correction layer: it never touches the backbone, the training data, or any gradient. Every correction is an edit to the memory bank $M$, so it applies to any nearest-neighbour detector with an explicit feature store.

Correction proceeds over $R$ rounds (Algorithm~\ref{alg:hitl}). In round $r$: (1) score all images against $M^{(r)}$; (2) form the error pool $E^{(r)} = \{x : \text{label}(x) \ne \text{predict}(x, M^{(r)}, \tau^{(r)})\}$; (3) present the expert one image from $E^{(r)}$; (4) the expert gives one binary label, ``normal'' or ``defect''; (5) apply the corresponding correction to obtain $M^{(r+1)}$. The loop stops at round $R$ or when the error pool is exhausted. The entire human contribution per round is one label on one image.

\subsubsection{False Positive Correction: Memory Bank Expansion}

A false positive occurs when a normal image $x_{\text{FP}}$ receives an anomaly score $s(x_{\text{FP}}) \ge \tau$, indicating that its patch features are anomalously distant from the current memory bank. The geometric interpretation is that $x_{\text{FP}}$ lies in a region of normal appearance space that is underrepresented in $M$: a gap in the normal manifold that causes the nearest-neighbour distance to exceed the anomaly threshold despite the image being normal.

The correction adds all 784 patch features of $x_{\text{FP}}$ directly to the memory bank, filling this gap:

\begin{equation}
M^{(r+1)} \leftarrow M^{(r)} \cup P(x_{\text{FP}})
\label{eq:fp_correction}
\end{equation}

This operation increases $|M|$ by 784 entries per FP correction. Without a size constraint, repeated FP corrections would cause the memory bank to grow without bound, increasing nearest-neighbour query time quadratically. To prevent this, we enforce a maximum bank size $N_{\max}$ via random subsampling after each FP correction:

\begin{equation}
M^{(r+1)} \leftarrow \text{RandomSample}\big(M^{(r)} \cup P(x_{\text{FP}}),\, N_{\max}\big)
\label{eq:fp_subsample}
\end{equation}

\noindent applied whenever the insertion would leave $|M^{(r+1)}| > N_{\max}$.

The value of $N_{\max}$ is set uniformly across categories, as detailed in Section~\ref{sec:implementation}. The random subsampling preserves the global coverage properties of the coreset while bounding computational cost; a control experiment with the cap removed (Section~\ref{sec:implementation}) shows it is not a sensitive parameter.

\subsubsection{Novelty-Gated Insertion}
\label{sec:gating}

\begin{figure}[!t]
\centering
\includegraphics[width=\columnwidth]{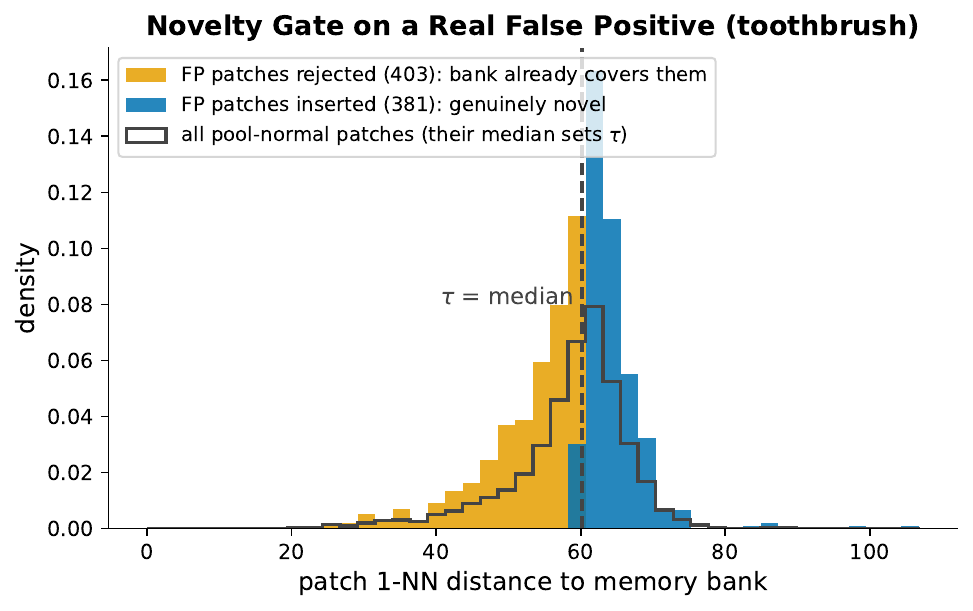}
\caption{The novelty gate on a real false positive (toothbrush, seed 10). The threshold $\tau_{\text{nov}}$ is the median 1-NN distance of all pool-normal patches to the baseline bank (black outline); of the flagged image's 784 patches, those within existing coverage (left of $\tau$) are rejected and only the genuinely novel ones (right of $\tau$) are inserted.}
\label{fig:novelty_gate}
\end{figure}

Adding all 784 patches of a corrected image is maximally effective when the bank genuinely undersamples normal appearance. On categories whose banks already cover the normal manifold well, however, most inserted patches duplicate existing coverage, and this redundant insertion can perturb previously correct decisions (Section~\ref{sec:cable_analysis}). The framework therefore supports gating each candidate patch on novelty: a patch $p \in P(x_{\text{FP}})$ is inserted only if its nearest-neighbour distance to the current bank exceeds a threshold $\tau_{\text{nov}}$,

\begin{equation}
M^{(r+1)} \leftarrow M^{(r)} \cup \big\{\, p \in P(x_{\text{FP}}) : \min_{m \in M^{(r)}} \lVert p - m \rVert_2 > \tau_{\text{nov}} \,\big\},
\label{eq:gating}
\end{equation}

where $\tau_{\text{nov}}$ is set once, before correction begins, as the median nearest-neighbour distance of all review-pool normal patches to the baseline bank (pool data only, never held-out images). Figure~\ref{fig:novelty_gate} shows the gate on a real toothbrush false positive: the half of its patches within existing coverage is rejected, the genuinely novel half inserted. The gate is self-calibrating (permissive on undersampled banks, conservative on well-covered ones) and needs no tuning: outcomes change smoothly between the 0.25 and 0.75 quantiles (Section~\ref{sec:gating_results}). The criterion has close relatives: it is an online, threshold-based analogue of the greedy $k$-center coreset rule that built the bank \cite{roth2022patchcore,sener2018coreset}, asking of each candidate patch at insertion time the coverage question that coreset construction asks of the training set; and a quantile-thresholded 1-NN novelty filter has recently been used to keep a time-series memory bank non-redundant under unsupervised stream adaptation \cite{han2025timerep}. What is new is the role: applied to human-verified corrections, the filter is not an economy measure. Section~\ref{sec:gating_results} shows it is the difference between a loop that harms well-covered categories and one that is safe to leave on. We evaluate the loop with and without the gate; gating is the recommended deployment configuration.

\subsubsection{False Negative Correction: Memory Bank Surgery}

A false negative occurs when an anomalous image $x_{\text{FN}}$ scores $s(x_{\text{FN}}) < \tau$: typically one or a few defect patches sit spuriously close to normal bank entries, inside the normal manifold.

The correction first identifies the anchor patch: the patch of $x_{\text{FN}}$ lying furthest from the current bank under the \emph{1-NN} distance,

\begin{equation}
p^* = \operatorname*{argmax}_{p \in P(x_{\text{FN}})} \; \min_{m \in M} \; \lVert p - m \rVert_2
\label{eq:anchor_patch}
\end{equation}

Note that Eq.~\ref{eq:anchor_patch} selects the anchor by 1-NN distance, whereas the image score of Eq.~\ref{eq:anomaly_score} is driven by the mean-of-nine statistic $d_9$. The two criteria do not in general identify the same patch, so removal is aimed at the most locally isolated patch rather than at the patch that actually set the score; Section~\ref{sec:fp_fn_ablation} shows this to be one of the two reasons FN correction behaves as a near no-op. Because $x_{\text{FN}}$ was classified normal, $p^*$ sits in a region of feature space that is ambiguously close to the normal manifold. The correction removes the $k_{\text{remove}} = 5$ nearest memory bank entries to $p^*$, creating a gap in the normal manifold at precisely the location of the anomalous patch:

\begin{align}
I_{\text{remove}} &= k\text{NN}(p^*, M^{(r)}, k=5) \label{eq:knn_remove} \\
M^{(r+1)} &\leftarrow M^{(r)} \setminus \{M^{(r)}_i : i \in I_{\text{remove}}\} \label{eq:fn_correction}
\end{align}

The excised gap raises future scores for patches falling in the same region, so similar defects are flagged; $k_{\text{remove}}=5$ balances gap size against over-removal. After each correction (FP or FN) the bank index is rebuilt and all images rescored.

\begin{algorithm}[!t]
\caption{HITL Memory Bank Correction}
\label{alg:hitl}
\begin{algorithmic}[1]
\Require $M$ (memory bank), $X_T$ (test images + labels), $R$ (rounds), strategy $\in \{$passive, active$\}$
\Ensure $M^{(R)}$ (corrected bank), \texttt{auroc\_curve}, AUCC
\State scores $\leftarrow$ \Call{ScoreAll}{$X_T$, $M$}
\State $\text{AUROC}_0 \leftarrow \text{AUROC}(\text{scores}, \text{labels}(X_T))$
\State corrected $\leftarrow \emptyset$; \texttt{auroc\_curve} $\leftarrow [\text{AUROC}_0]$
\For{$r = 1, 2, \ldots, R$}
    \State scores$^{(r)} \leftarrow$ \Call{ScoreAll}{$X_T$, $M$}
    \State $\tau^{(r)} \leftarrow \text{percentile}_{90}(\text{scores}^{(r)}[\text{normal}])$
    \State $E^{(r)} \leftarrow \{x : \text{label}(x) \ne \text{predict}(x, M, \tau^{(r)})\} \setminus \text{corrected}$
    \If{$E^{(r)} = \emptyset$}
        \State \textbf{break} \Comment{no errors remain}
    \EndIf
    \State $x^* \leftarrow$ \Call{QueryStrategy}{$E^{(r)}$, strategy, $\tau^{(r)}$} \Comment{Eq.~\ref{eq:passive}/\ref{eq:active}}
    \State \textbf{[Human]} inspects $x^*$; labels it \textit{normal} or \textit{defective} \Comment{single binary decision}
    \If{$x^*$ is FP (human says \textit{normal}, model said defective)}
        \State $M \leftarrow$ \Call{FPCorrect}{$x^*$, $M$, $N_{\max}$}
    \Else \Comment{FN: human says \textit{defective}, model said normal}
        \State $M \leftarrow$ \Call{FNCorrect}{$x^*$, $M$, $k_{\text{remove}}=5$}
    \EndIf
    \State corrected $\leftarrow$ corrected $\cup \{x^*\}$
    \State $\text{AUROC}_r \leftarrow \text{AUROC}(\Call{ScoreAll}{X_T, M}, \text{labels})$
    \State \texttt{auroc\_curve.append}($\text{AUROC}_r$)
\EndFor
\State $g_r \leftarrow \max(0, \text{AUROC}_r - \text{AUROC}_0)$, $r = 1, \ldots, R$
\State $\text{AUCC} \leftarrow \dfrac{1}{R(1-\text{AUROC}_0)} \sum_r g_r$
\State \Return $M$, \texttt{auroc\_curve}, AUCC
\end{algorithmic}
\end{algorithm}

\subsection{Querying Strategies}

The query strategy determines only the \emph{order} in which errors are reviewed, not the correction mechanism. We evaluate two:

\textbf{Passive querying} draws uniformly at random from the error pool,

\begin{equation}
x^* \sim \text{Uniform}(E^{(r)}),
\label{eq:passive}
\end{equation}

\noindent requiring no computation beyond the scoring pass and sampling FP and FN errors in proportion to their prevalence \cite{settles2010active}.

\textbf{Active querying} implements uncertainty sampling \cite{settles2010active,lewis1994sequential}, selecting the image whose score is closest to the decision threshold:

\begin{equation}
x^* = \operatorname*{argmin}_{x \in E^{(r)}} \; \lvert s(x) - \tau^{(r)} \rvert.
\label{eq:active}
\end{equation}

\noindent Its rationale, that boundary-adjacent samples are most informative, assumes local model updates. Section~\ref{sec:results} shows empirically that this assumption fails for memory bank correction, whose updates are global (Section~\ref{sec:discussion}).

\subsection{Evaluation Metric: Area Under the Correction Curve (AUCC)}

AUROC before and after correction misses the \emph{efficiency} of the process: how quickly improvement arrives over the review budget. The \textbf{Area Under the Correction Curve (AUCC)} captures it. With $\text{AUROC}_r$ the AUROC after round $r$ ($r=0$ the baseline), define the per-round gain $g_r = \max(0,\, \text{AUROC}_r - \text{AUROC}_0)$ and

\begin{equation}
\text{AUCC} = \frac{1}{R\,(1-\text{AUROC}_0)} \sum_{r=1}^{R} g_r \;\in\; [0, 1],
\label{eq:aucc}
\end{equation}

\noindent i.e.\ the cumulative gain normalised by the area of immediate, sustained perfect correction. AUCC is comparable across baselines and rewards \emph{early} improvement: the same final AUROC scores higher if reached in fewer reviews, matching the practical preference for fewer human interventions. It is an AUROC-specific instance of the area-under-the-learning-curve (ALC) family standard in active-learning evaluation \cite{guyon2011alc}, adapted to correction: gains are measured against an already-deployed baseline and normalised by the remaining headroom $1-\text{AUROC}_0$, so scores are comparable across categories with different starting points.

\subsection{Experimental Setup}

\subsubsection{Dataset}

All experiments use MVTec AD \cite{bergmann2019mvtec}, the standard industrial anomaly detection benchmark: 5{,}354 images across 15 categories (5 textures, 10 objects) spanning 73 defect types, from fixed-pose rigid objects (bottle) to deformable configurations (cable) and random textures (leather, wood). Training sets contain only defect-free images; test sets contain both classes.

\subsubsection{Category Selection: Stratified by Separation Index}

\begin{figure}[!t]
\centering
\includegraphics[width=\columnwidth]{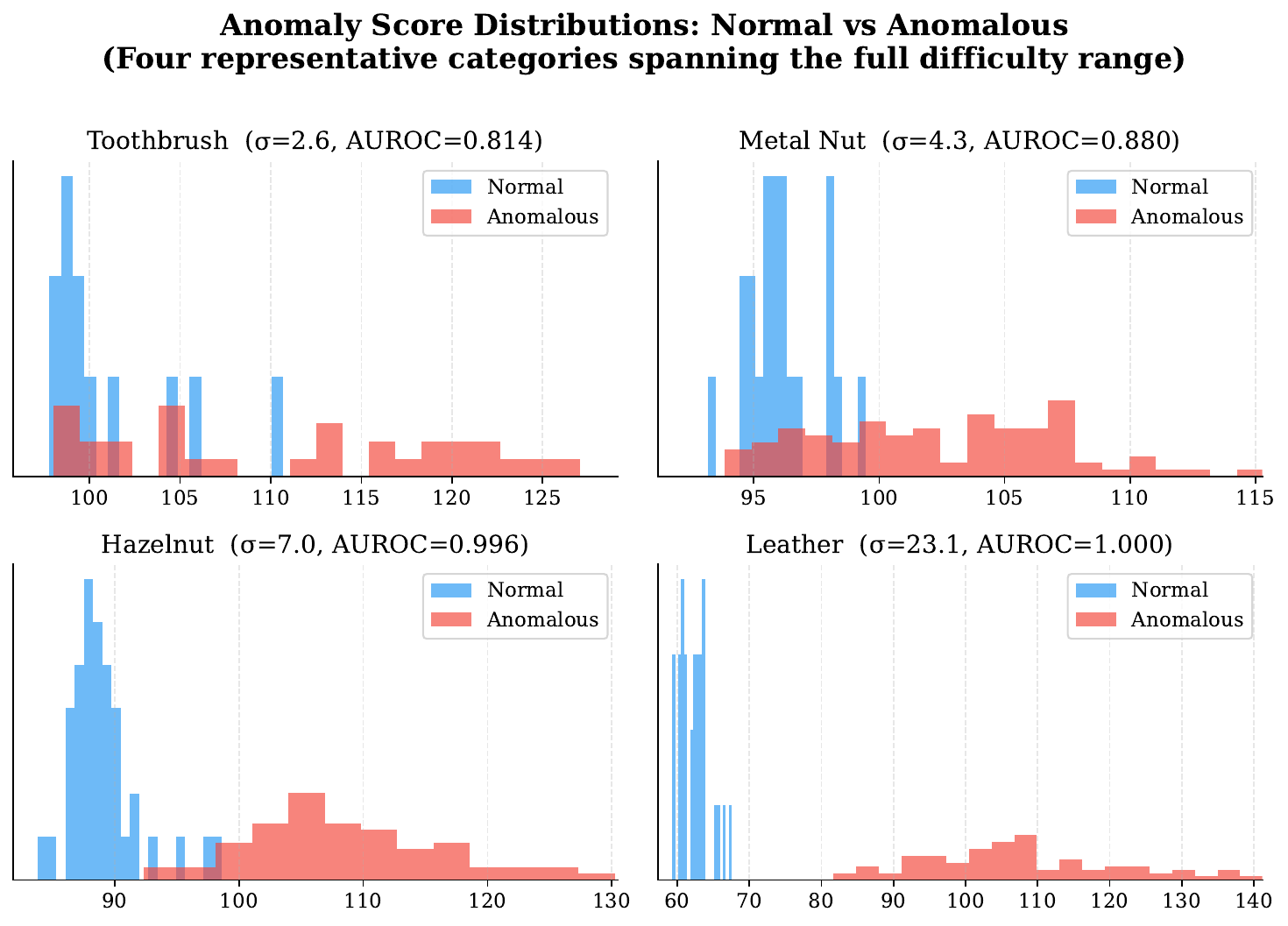}
\caption{Anomaly score distributions for four representative categories spanning the full difficulty range (blue = normal, red = anomalous). Overlap decreases as $\sigma$ increases, from heavily overlapping (toothbrush) to fully separated (leather).}
\label{fig:score_distributions}
\end{figure}

We evaluate all 15 MVTec AD categories using passive and active querying strategies. The separation index $\sigma$, defined as the normalised distance between normal and anomalous score distributions, predicts correction utility and is used to interpret results across categories:

\begin{equation}
\sigma = \frac{\mu_{\text{anomalous}} - \mu_{\text{normal}}}{\sigma_{\text{normal}}}
\label{eq:separation}
\end{equation}

where $\mu_{\text{anomalous}}$ and $\mu_{\text{normal}}$ denote the mean anomaly score for each class, and $\sigma_{\text{normal}}$ denotes the standard deviation of the normal class score distribution. Normalising by $\sigma_{\text{normal}}$ (rather than a pooled variance) reflects how many normal-distribution standard deviations separate the two class means, a natural unit given that PatchCore's classification threshold itself is set from the normal score distribution. A high separation index indicates well-separated score distributions (easy category); a low index indicates overlapping distributions (hard category), as illustrated in Fig.~\ref{fig:score_distributions}. Table~\ref{tab:categories} lists all 15 categories with their separation indices, baseline AUROCs and dataset statistics; the baselines average 0.9342 and span 0.8139 (toothbrush) to 1.0000 (bottle, leather).

\begin{table}[!t]
\caption{All 15 MVTec AD categories, ordered by separation index $\sigma$ (descending).}
\label{tab:categories}
\centering
\begin{tabular}{lccccc}
\toprule
\textbf{Category} & \textbf{Type} & $\boldsymbol{\sigma}$ & \textbf{Base.} & \textbf{Train} & \textbf{Test (g/a)} \\
\midrule
Bottle     & Object  & 50.5 & 1.0000 & 209 & 20/63 \\
Leather    & Texture & 23.1 & 1.0000 & 245 & 32/92 \\
Carpet     & Texture & 9.5  & 0.9932 & 280 & 28/89 \\
Wood       & Texture & 7.6  & 0.9912 & 247 & 19/60 \\
Hazelnut   & Object  & 7.0  & 0.9964 & 391 & 40/70 \\
Tile       & Texture & 6.0  & 0.9776 & 230 & 33/84 \\
Zipper     & Object  & 4.9  & 0.8787 & 240 & 32/119 \\
Cable      & Object  & 4.6  & 0.9185 & 224 & 58/92 \\
Capsule    & Object  & 4.3  & 0.9290 & 219 & 23/109 \\
Pill       & Object  & 4.3  & 0.9242 & 267 & 26/141 \\
Metal Nut  & Object  & 4.3  & 0.8803 & 220 & 22/93 \\
Transistor & Object  & 3.7  & 0.9558 & 213 & 60/40 \\
Screw      & Object  & 3.0  & 0.8703 & 320 & 41/119 \\
Toothbrush & Object  & 2.6  & 0.8139 & 60  & 12/30 \\
Grid       & Texture & 2.6  & 0.8839 & 264 & 21/57 \\
\bottomrule
\end{tabular}
\end{table}

\subsubsection{Implementation Details and Bank Size Caps}
\label{sec:implementation}

The WideResNet-50 backbone uses ImageNet-pretrained weights loaded via the \textit{timm} library (version 0.9+) and is kept frozen throughout. Input images are resized to 224$\times$224 pixels and normalised with ImageNet mean $[0.485, 0.456, 0.406]$ and standard deviation $[0.229, 0.224, 0.225]$. Features are extracted from backbone stages 2 and 3 using the \texttt{features\_only} interface. The memory bank nearest-neighbour index uses $k=9$, Euclidean distance, and a ball-tree data structure (scikit-learn \texttt{NearestNeighbors} with \texttt{algorithm='ball\_tree'}, \texttt{n\_jobs=-1}) for efficient batch querying.

The memory bank is capped at $N_{\max} = 12{,}000$ patches, uniformly across all categories: when an insertion would exceed the cap, the bank is randomly subsampled back to $N_{\max}$. The cap exists only to bound scoring cost; it is not a sensitive parameter. A 20-seed control run with the cap removed entirely changes no category's mean held-out delta except transistor ($+0.021$, the only Holm-significant shift) and doubles the scoring cost.

Each HITL experiment runs for $R = 30$ correction rounds per strategy per category. The ball-tree index is rebuilt after each correction to reflect the updated bank.

\subsubsection{Held-Out Evaluation Protocol}
\label{sec:heldout_protocol}

A methodological subtlety of memory bank correction is that Eq.~\ref{eq:fp_correction} adds the corrected image's \emph{own} patches to the bank. If the same images are used both for correction and for evaluation, AUROC trends toward 1.0 partly because corrected normal images are subsequently scored against exact copies of their own features. That is memorisation, not generalisation. All headline results in this paper therefore use a held-out protocol.

For each experimental run, the category's test images are split 70/30 (stratified by label) into a \emph{review pool} and a \emph{held-out set}. Only pool images may enter the error pool $E^{(r)}$ and receive corrections; the decision threshold $\tau^{(r)}$ is computed from pool normals only. The held-out set is rescored against the updated bank after every correction round but never receives corrections and never influences querying or thresholding. Held-out AUROC after round $r$ is the generalisation measure: it reflects whether corrections improve detection on images the loop has never seen. AUCC (Eq.~\ref{eq:aucc}) is likewise computed on the held-out curve. Pool AUROC is reported only as a secondary, self-consistency measure and is explicitly marked as inflated by memorisation wherever it appears.

Every (category, strategy) condition is run at 20 independent split seeds (10--29), each seed drawing a different pool/held-out partition and, for passive querying, a different random review order. Reported values are means over the 20 seeds with standard deviations; per-seed distributions are reported in Section~\ref{sec:multiseed}. Statistical significance of per-category improvement is assessed with a two-sided Wilcoxon signed-rank test on the 20 paired per-seed held-out deltas (final minus baseline), and passive-vs-active differences with the same test on paired per-seed endpoints; both are Holm-corrected across the 15 categories at $\alpha = 0.05$. Seeds 0--9 were reserved for earlier exploratory work and are excluded from all reported results. A uniform bank cap $N_{\max}=12{,}000$ is used for all categories under this protocol (Section~\ref{sec:implementation}).

\subsubsection{Cold-Start Protocol}
\label{sec:coldstart_protocol}

The protocol above corrects a fully trained bank. To evaluate the deployment scenario where little training data exists (a new production line with only 10--20 verified ``golden'' samples), we additionally run a cold-start protocol. For each seed, the initial bank is built from only $k \in \{10, 20\}$ training normals drawn with a seed-coupled random number generator: their patch features are extracted with the same backbone and reduced by greedy $k$-center coreset selection at the main protocol's 1\% rate (78 patches at $k=10$, 156 at $k=20$). The HITL loop then runs unchanged, with the same 70/30 pool/held-out splits as the full-training runs on matched seeds, and the budget extended to 90 rounds to accommodate the larger initial error pools. Because splits are seed-matched, the full-training result on the same seed is a paired reference: we report the \emph{recovery fraction}, i.e.\ the fraction of the cold-start-to-full-training held-out gap closed by correction, and the median number of reviews to reach parity with the full-training bank.

All experiments are conducted on NVIDIA GPUs; timing details are reported in Section~\ref{sec:limitations}.

\section{Experimental Results}
\label{sec:results}

We present the full 15-category benchmark on fully trained banks, then the effect of the novelty gate (Section~\ref{sec:gating_results}) and the cold-start setting (Section~\ref{sec:coldstart_results}), followed by the analyses that explain the results: key categories, strategy equivalence, seed robustness, label economics, and the FP/FN asymmetry.

\subsection{Full 15-Category Results Overview}

Table~\ref{tab:full_results} reports held-out results for all 15 MVTec AD categories under the protocol of Section~\ref{sec:heldout_protocol}: means over 20 pool/held-out splits (seeds 10--29), ordered by baseline AUROC; ``HO'' is the held-out set, which never receives corrections.

\begin{table*}[!t]
\caption{Complete 15-category HITL results under the held-out protocol (means over 20 seeds). HO = held-out set, never corrected. Bold improvements are consistent across seeds ($\ge$15/20 seeds improved); negative improvements are shown as-is; the ungated correction loop is not beneficial on every category. ``Gated $\Delta$'' is the passive loop with novelty-gated insertion (Section~\ref{sec:gating}), the recommended deployment configuration. $^{*}$ marks a $\Delta$ significantly different from zero (two-sided Wilcoxon signed-rank on per-seed deltas, Holm-corrected across 15 categories, $\alpha=0.05$); it is reported for the passive and gated configurations. Active deltas are not separately tested against zero, because no category shows a significant passive-vs-active difference under the same test.}
\label{tab:full_results}
\centering
\begin{tabular}{lcccccccc}
\toprule
\textbf{Category} & \textbf{Type} & \textbf{Sep.($\sigma$)} & \textbf{Full-test Base.} & \textbf{HO Base.} & \textbf{Pass.\ HO Final} & \textbf{Pass.\ $\Delta$} & \textbf{Act.\ $\Delta$} & \textbf{Gated $\Delta$} \\
\midrule
Toothbrush & Object  & 2.6  & 0.8139 & 0.8264 & 0.9722 & $\mathbf{+0.146}^{*}$ & $\mathbf{+0.146}$ & $\mathbf{+0.101}^{*}$ \\
Grid       & Texture & 2.6  & 0.8839 & 0.8662 & 0.8190 & $-0.047^{*}$ & $-0.056$ & $-0.031^{*}$ \\
Screw      & Object  & 3.0  & 0.8703 & 0.8703 & 0.9105 & $\mathbf{+0.040}^{*}$ & $\mathbf{+0.043}$ & $\mathbf{+0.048}^{*}$ \\
Transistor & Object  & 3.7  & 0.9558 & 0.9623 & 0.9379 & $-0.024^{*}$ & $-0.031$ & $+0.005$ \\
Metal Nut  & Object  & 4.3  & 0.8803 & 0.9000 & 0.9980 & $\mathbf{+0.098}^{*}$ & $\mathbf{+0.098}$ & $\mathbf{+0.089}^{*}$ \\
Pill       & Object  & 4.3  & 0.9242 & 0.9283 & 0.9051 & $-0.023^{*}$ & $-0.026$ & $+0.007$ \\
Capsule    & Object  & 4.3  & 0.9290 & 0.9403 & 0.9310 & $-0.009$ & $-0.012$ & $+0.003$ \\
Cable      & Object  & 4.6  & 0.9185 & 0.9260 & 0.9069 & $-0.019$ & $-0.018$ & $+0.006$ \\
Zipper     & Object  & 4.9  & 0.8787 & 0.8772 & 0.9317 & $\mathbf{+0.054}^{*}$ & $\mathbf{+0.054}$ & $\mathbf{+0.051}^{*}$ \\
Tile       & Texture & 6.0  & 0.9776 & 0.9806 & 0.9679 & $-0.013^{*}$ & $-0.012$ & $-0.001$ \\
Hazelnut   & Object  & 7.0  & 0.9964 & 0.9960 & 0.9985 & $+0.002$ & $+0.002$ & $+0.003$ \\
Wood       & Texture & 7.6  & 0.9912 & 0.9898 & 0.9940 & $+0.004$ & $+0.004$ & $+0.004$ \\
Carpet     & Texture & 9.5  & 0.9932 & 0.9930 & 0.9961 & $+0.003$ & $+0.003$ & $+0.003$ \\
Leather    & Texture & 23.1 & 1.0000 & 1.0000 & 1.0000 & $0.000$ & $0.000$ & $0.000$ \\
Bottle     & Object  & 50.5 & 1.0000 & 1.0000 & 0.9991 & $-0.001$ & $-0.001$ & $-0.000$ \\
\bottomrule
\end{tabular}
\end{table*}

The benchmark partitions into three regimes. \textbf{Four categories gain $+0.040$ to $+0.146$ AUROC, on 15--20 of 20 seeds each}: toothbrush ($+0.146$, 19/20), metal nut ($+0.098$, 20/20), zipper ($+0.054$, 20/20), screw ($+0.040$, 15/20). These are exactly the categories whose baseline errors are false positives from an undersampled normal manifold, which FP corrections repair. \textbf{Five near-ceiling categories} (baseline $\ge 0.99$) are unaffected. \textbf{Six categories are flat-to-negative}, losing up to $0.047$ AUROC (grid, transistor, pill, capsule, cable, tile): their pool AUROC still climbs (the memorisation artefact), but most inserted patches duplicate existing coverage, and this redundant insertion degrades held-out performance. A 20-seed control with the bank cap removed leaves these categories unchanged, ruling out subsampling churn as the cause; Section~\ref{sec:gating_results} shows the novelty gate eliminates the degradation on five of the six.

These regimes survive formal testing. Applying a two-sided Wilcoxon signed-rank test to the 20 per-seed held-out deltas of each category (Holm-corrected across the 15 categories), all four winners are significantly positive (toothbrush, metal nut, zipper: adjusted $p < 10^{-4}$; screw: adjusted $p = 0.032$), and four of the six flat-to-negative categories are significantly \emph{negative} (grid, pill, tile, transistor; adjusted $p \le 0.032$), which confirms that the degradation on those categories is systematic rather than noise. Cable and capsule are statistically indistinguishable from zero (adjusted $p = 0.057$ and $0.28$). The same paired test applied to passive-vs-active per-seed differences finds no significant difference on any category (all unadjusted $p \ge 0.068$; all adjusted $p = 1$), formalising the strategy-equivalence claim of Section~\ref{sec:strategy_comparison}.

\subsection{Novelty-Gated Insertion Eliminates the Harm}
\label{sec:gating_results}

The novelty gate (Section~\ref{sec:gating}) changes the risk profile of the entire benchmark. With gating enabled (final column of Table~\ref{tab:full_results}), every flat-to-negative category except grid turns statistically neutral or positive: transistor $-0.024 \rightarrow +0.005$, pill $-0.023 \rightarrow +0.007$, cable $-0.019 \rightarrow +0.006$, tile $-0.013 \rightarrow -0.001$, capsule $-0.009 \rightarrow +0.003$; the paired per-seed improvement over the ungated loop is Holm-significant on all five. All four winners remain Holm-significantly positive (toothbrush $+0.101$, metal nut $+0.089$, zipper $+0.051$, screw $+0.048$); the price of conservatism is a modest reduction on toothbrush ($-0.044$ vs.\ ungated, not Holm-significant across seeds) and metal nut ($-0.009$, Holm-significant but an order of magnitude smaller than the retained gain). The result is a deployment configuration in which \textbf{no category except grid is significantly harmed, while every previous winner remains a winner}. Figure~\ref{fig:coldstart_curves}(d) shows the effect directly on pill: the ungated trajectory drifts below baseline round by round, while the gated trajectory stays above it.

The gate's insertion rates show the self-calibration at work. On well-covered banks it makes corrections conservative (tile inserts a mean of only 65 of 784 candidate patches per FP correction, transistor 78) and the harm of redundant insertion vanishes; on undersampled banks it stays permissive (metal nut inserts 271, zipper 236) and the gains survive. Grid is the diagnostic exception: it inserts 230 patches per correction, more than metal nut, yet still degrades, so its errors are not a coverage problem at all; we return to it in Section~\ref{sec:cable_analysis}. The threshold requires no tuning: repeating the full sweep with $\tau_{\text{nov}}$ at the 0.25 and 0.75 quantiles of the same distance distribution shifts outcomes smoothly and modestly. At 0.25 the rescue of the negative categories weakens (of the four significantly negative categories, only transistor reaches non-negative); at 0.75 it strengthens (grid improves to $-0.019$) but a real cost appears on metal nut ($-0.043$ vs.\ ungated, Holm-significant, 0/20 seeds better). The median sits at the sweet spot, and no quantile exhibits knife-edge behaviour.

\subsection{Cold-Start Deployment: Near-Full Performance from Ten Samples}
\label{sec:coldstart_results}

\begin{figure}[!t]
\centering
\includegraphics[width=\columnwidth]{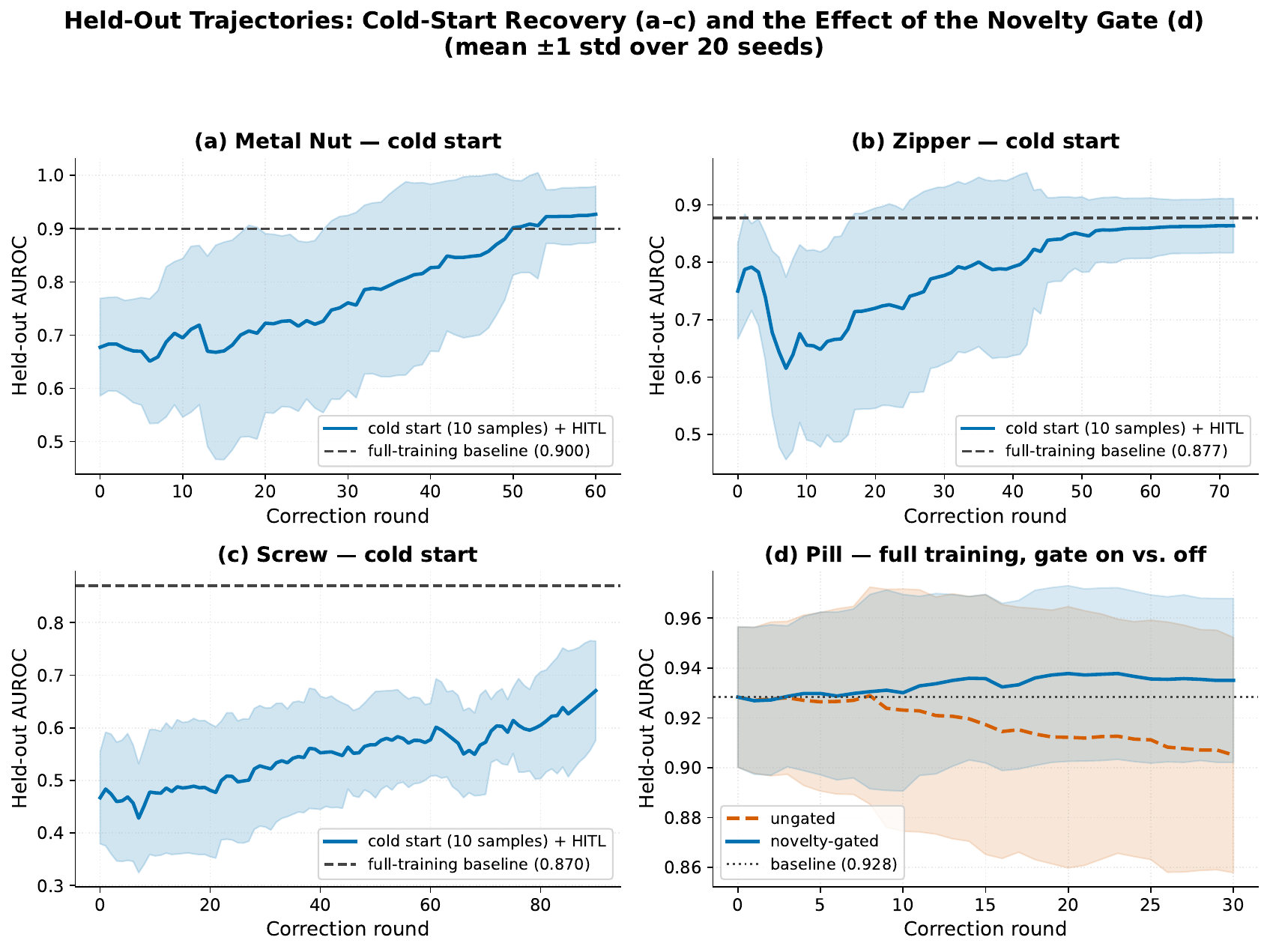}
\caption{Held-out trajectories (mean $\pm 1$ std over 20 seeds). (a--c) Cold-start recovery from 10 golden samples with the novelty gate: metal nut crosses its full-training baseline around round 50 and ends above it; zipper converges to just below parity after a transient early dip (corrections on a 78-patch bank briefly disturb scores before coverage accumulates); screw is still climbing when the 90-round budget ends, far from parity (the budget-limited case). (d) Pill on a fully trained bank: the ungated loop drifts below baseline while the novelty-gated loop stays above it, showing the gate's effect in a single panel.}
\label{fig:coldstart_curves}
\end{figure}

Deployment-time correction delivers its largest gains where training data is scarcest. Table~\ref{tab:coldstart} reports the cold-start protocol (Section~\ref{sec:coldstart_protocol}) with $k=10$ golden samples and the novelty gate enabled. Starting from banks of just 78 patches, the correction loop produces Holm-significant held-out improvements on \textbf{12 of 15 categories, with no category significantly harmed}, and recovers on average 80\% of the gap to the full-training bank (median 0.66 over the 11 categories with a meaningful gap). On two categories the corrected cold-start bank ends substantially \emph{above} the full-training reference: metal nut ($0.927$ vs.\ $0.900$) and toothbrush ($0.904$ vs.\ $0.826$). Carpet and hazelnut also finish fractionally above their references ($\le 0.003$), but both start within $0.01$ of ceiling, where the margin carries no weight. Operator-curated insertion can therefore beat exhaustive training data on quality as well as on labelling cost. Hazelnut, bottle, zipper and carpet reach or exceed parity within a median of 4--19 reviews (medians over the seeds that reach parity; Table~\ref{tab:coldstart} reports how many of the 20 do). Figure~\ref{fig:coldstart_curves}(a--c) shows the three characteristic trajectories: crossing the full-training baseline (metal nut), converging to parity (zipper), and still climbing when the budget ends (screw).

\begin{table*}[!t]
\caption{Cold-start results: initial bank built from only $k=10$ training normals (78 patches), passive querying with novelty-gated insertion, 90-round budget, means over 20 seeds. ``Full Ref.''\ = full-training-bank held-out baseline on the same seed (paired). Recovery = fraction of the cold-to-full gap closed (shown where the gap exceeds 0.02). ``Labels to parity'' = median reviews to reach the full-training reference, with the number of seeds reaching it. $^{*}$ = Holm-significant $\Delta$ (Wilcoxon, $\alpha=0.05$); no category is significantly negative.}
\label{tab:coldstart}
\centering
\begin{tabular}{lccccccc}
\toprule
\textbf{Category} & \textbf{Cold Base} & \textbf{Post-HITL} & \textbf{Full Ref.} & $\boldsymbol{\Delta}$ & \textbf{Recovery} & \textbf{Reviews} & \textbf{Labels to Parity} \\
\midrule
Bottle     & 0.957 & 0.996 & 1.000 & $+0.039^{*}$ & 0.92 & 14 & 5 (7/20) \\
Cable      & 0.756 & 0.866 & 0.926 & $+0.110^{*}$ & 0.65 & 50 & 35 (1/20) \\
Capsule    & 0.560 & 0.699 & 0.940 & $+0.139^{*}$ & 0.37 & 80 & --- (0/20) \\
Carpet     & 0.987 & 0.996 & 0.993 & $+0.008^{*}$ & --- & 18 & 8 (2/20) \\
Grid       & 0.681 & 0.719 & 0.866 & $+0.038$ & 0.20 & 37 & 24 (6/20) \\
Hazelnut   & 0.921 & 0.998 & 0.996 & $+0.077^{*}$ & 1.03 & 16 & 4 (14/20) \\
Leather    & 1.000 & 1.000 & 1.000 & $+0.000$ & --- & 19 & --- (0/20) \\
Metal Nut  & 0.677 & 0.927 & 0.900 & $+0.250^{*}$ & 1.12 & 42 & 36 (15/20) \\
Pill       & 0.579 & 0.791 & 0.928 & $+0.212^{*}$ & 0.61 & 77 & 22 (1/20) \\
Screw      & 0.467 & 0.671 & 0.870 & $+0.204^{*}$ & 0.50 & 90 & --- (0/20) \\
Tile       & 0.968 & 0.972 & 0.981 & $+0.004$ & --- & 26 & 4 (2/20) \\
Toothbrush & 0.732 & 0.904 & 0.826 & $+0.172^{*}$ & 1.82 & 12 & 8 (15/20) \\
Transistor & 0.656 & 0.857 & 0.962 & $+0.202^{*}$ & 0.66 & 32 & 23 (1/20) \\
Wood       & 0.982 & 0.989 & 0.990 & $+0.007^{*}$ & --- & 10 & --- (0/20) \\
Zipper     & 0.750 & 0.864 & 0.877 & $+0.114^{*}$ & 0.89 & 44 & 19 (9/20) \\
\bottomrule
\end{tabular}
\end{table*}

The cold-start result survives both protocol variations. Without the gate ($k=10$ ungated), 11 of 15 categories are Holm-significantly positive with mean recovery 0.89 and still no significant negatives; with $k=20$ (156 patches, higher cold baselines and smaller gaps), 8 of 15 remain significantly positive with mean recovery 0.85. Neither the choice of $k$ nor the gate produces the effect. Even grid, the open failure of the full-training setting, is neutral here ($+0.038$, not significant), and the flat-to-negative group of Table~\ref{tab:full_results} becomes uniformly positive: when the bank genuinely undersamples normal appearance, every correction teaches something. The honest limits: capsule recovers only 37\% of its gap, screw exhausts the 90-round budget at half recovery, and pill reaches 61\% after a mean of 77 reviews. On categories with large error pools the loop is label-efficient, not label-free.

\subsection{Key Category Analysis}

We analyse the two instructive extremes: toothbrush, the largest success, and the flat-to-negative group, the method's boundary.

\subsubsection{Toothbrush: Maximum Feedback Benefit}

Toothbrush is the hardest category in our evaluation (separation index $2.6\sigma$, baseline AUROC 0.8139) and produces the largest held-out improvement observed: $0.826 \rightarrow 0.972$ ($+0.146$) on images the correction loop never touches, improving on 19 of 20 splits (worst split $-0.028$) with a mean of only $\sim$13 corrections. Pool AUROC reaches 1.000 on every split, but per Section~\ref{sec:heldout_protocol} that figure partly reflects memorisation of corrected images and should not be quoted as the result.

\begin{figure*}[!t]
\centering
\includegraphics[width=\textwidth]{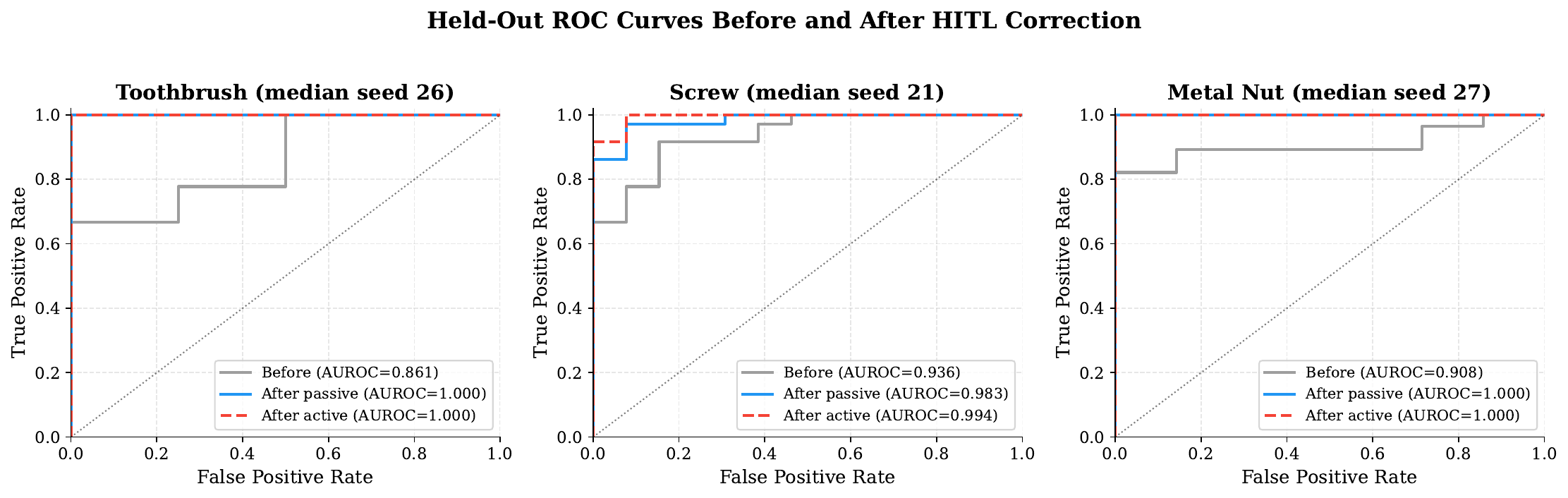}
\caption{Held-out ROC curves before (grey) and after correction under passive (blue) and active (red dashed) querying, for the three largest-improvement categories at their median-improvement seed. The corrected ROC dominates the baseline everywhere; passive and active endpoints are nearly identical.}
\label{fig:roc_before_after}
\end{figure*}

Figure~\ref{fig:roc_before_after} shows held-out ROC curves before and after correction at the median seeds: the corrected curves dominate at every operating point, so the gain is not an artefact of the AUROC summary. Why do so few reviews suffice? Clearing one falsely flagged image adds its 784 patches to a bank of only 470, which lowers nearest-neighbour distances for every image (held-out included) whose patches fall near the newly covered region. One label redefines a region of normal appearance; this transfer to held-out data is what separates correction from memorisation. The consequence is visible per round: on toothbrush and metal nut the entire gain arrives in one or two decisive FP corrections (on toothbrush a single round contributes $+0.139$) and most rounds contribute nothing, while screw, with a larger and noisier error pool, accumulates its gain gradually. The loop's value lies in reaching those images, not in correction volume. The FP/FN ablation (Section~\ref{sec:fp_fn_ablation}) and the matched-budget control (Section~\ref{sec:label_economics}) complete the mechanism account.

\subsubsection{Where Correction Fails: The Flat-to-Negative Group}
\label{sec:cable_analysis}

The six flat-to-negative categories share a profile: their banks are comparatively large, their remaining errors genuinely ambiguous, so FP corrections add patches the bank already covers, and the redundant insertion perturbs previously correct decisions. Grid is the clearest case ($-0.047$, degrading on 15/20 splits). The novelty gate (Section~\ref{sec:gating_results}) resolves five of the six by refusing exactly these redundant insertions. Grid alone remains significantly negative when gated ($-0.031$): it inserts 230 patches per correction yet still degrades, so its failure is not a coverage problem. It is the method's one open failure in the mature-bank setting, though it is unharmed at cold start (Table~\ref{tab:coldstart}). The practical rule: enable the gate by default and confirm per category with a held-out check.

\subsection{Strategy Comparison}
\label{sec:strategy_comparison}

\begin{figure}[!t]
\centering
\includegraphics[width=\columnwidth]{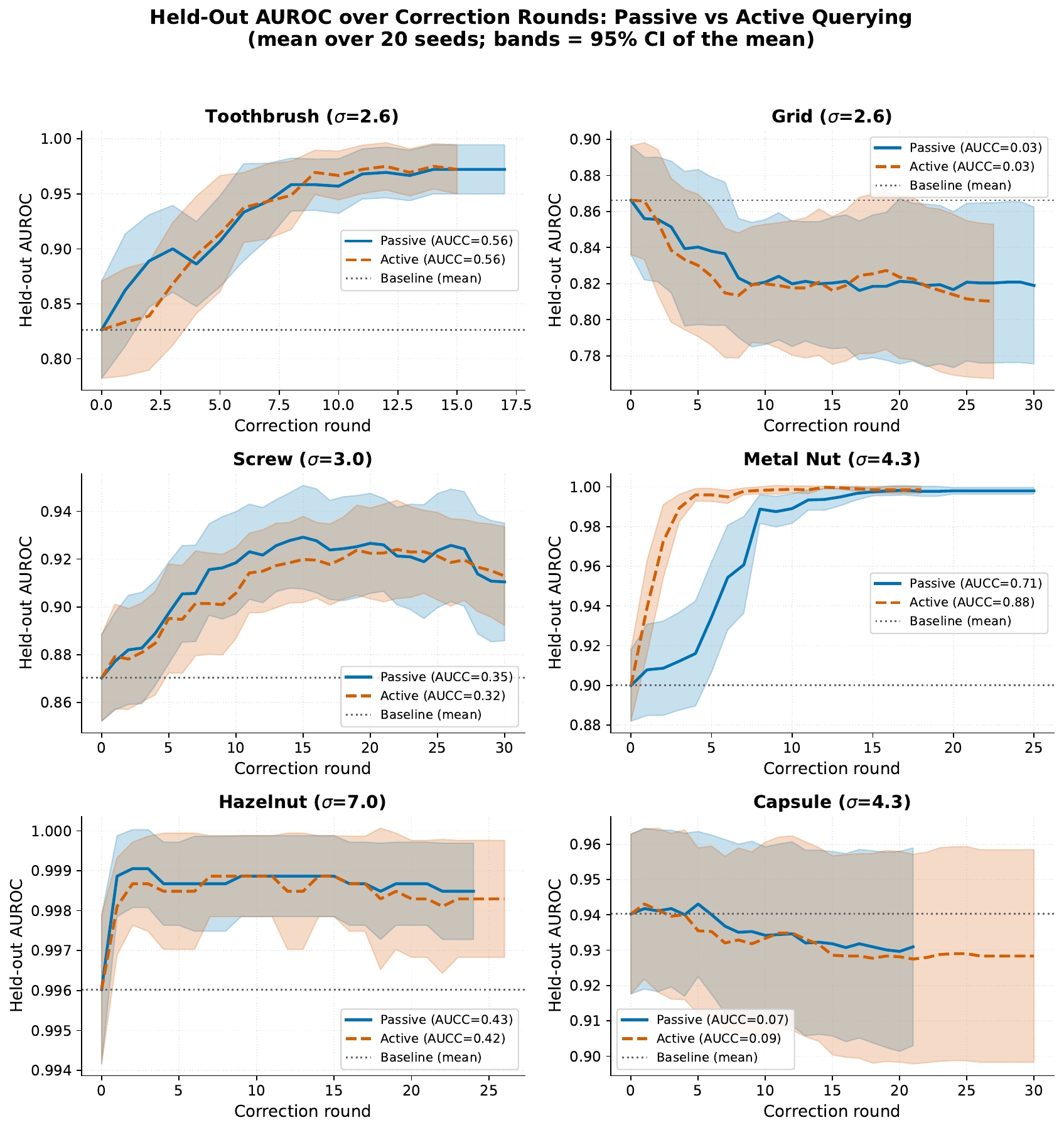}
\caption{Held-out AUROC over correction rounds for passive (blue) and active (red) querying, on six categories spanning the difficulty range (mean over 20 seeds; bands are the 95\% confidence interval of the mean, $t$-based with $n=20$, not the seed-to-seed spread). The bands overlap at every round except on metal nut, where active querying rises faster over the first few rounds before both strategies converge to the same endpoint: a difference in speed, not destination.}
\label{fig:strategy_curves}
\end{figure}

\textbf{Passive and active querying are statistically indistinguishable}: a paired Wilcoxon test on per-seed held-out endpoints finds no significant difference on any category, and mean improvements agree within $\pm 0.01$ AUROC on all 15 categories (Table~\ref{tab:full_results}); the residual differences sit well inside seed-to-seed variance, and the per-seed distributions of the two strategies overlap heavily on every category (Section~\ref{sec:multiseed}). Mean held-out trajectories with 95\% confidence bands overlap at every round on 14 of 15 categories (Fig.~\ref{fig:strategy_curves}). The one exception concerns speed rather than destination: on metal nut, active querying reaches the ceiling in 4 rounds versus 9 for passive (per-seed means; mean AUCC 0.88 vs.\ 0.71), the only stretch where the bands separate, yet both converge to the same endpoint. Note that time-to-ceiling is shorter than session length: metal nut sessions run a mean of $\sim$20 reviews, but the gain is complete well before the error pool is exhausted. An earlier single-seed finding that passive wins on 9 of 11 categories does not survive this evaluation, and we retract it; the mechanism (global bank updates plus a budget that exhausts the error pool under either ordering) is discussed in Section~\ref{sec:discussion}.

Held-out AUCC follows the same pattern as the AUROC deltas: correction utility concentrates in the low-$\sigma$ categories with FP-dominated error pools (Table~\ref{tab:full_results}), and passive and active AUCC agree on every category except metal nut.

\subsection{Ablation Study: $k_{\text{remove}}$ Parameter}

Ablating $k_{\text{remove}}$, the number of bank entries removed per FN correction, on toothbrush across $k \in \{1, 3, 5, 9\}$ gives identical results in every case (AUCC 0.9984, final AUROC 1.0000), because that session drew only FP corrections and the parameter never triggered. \textit{(Single seed 42, pool-only evaluation, predating the held-out protocol; the conclusion concerns the correction mechanism, not the split.)} Toothbrush sessions are not generally FN-free: under the held-out protocol they average 5.0 FN corrections (Table~\ref{tab:fp_fn_ablation}), which Section~\ref{sec:fp_fn_ablation} shows to be near no-ops. We use $k=5$ throughout, large enough to open a gap in the manifold and small enough to avoid over-removal.

\subsection{Multi-Seed Validation}
\label{sec:multiseed}

\begin{figure}[!t]
\centering
\includegraphics[width=\columnwidth]{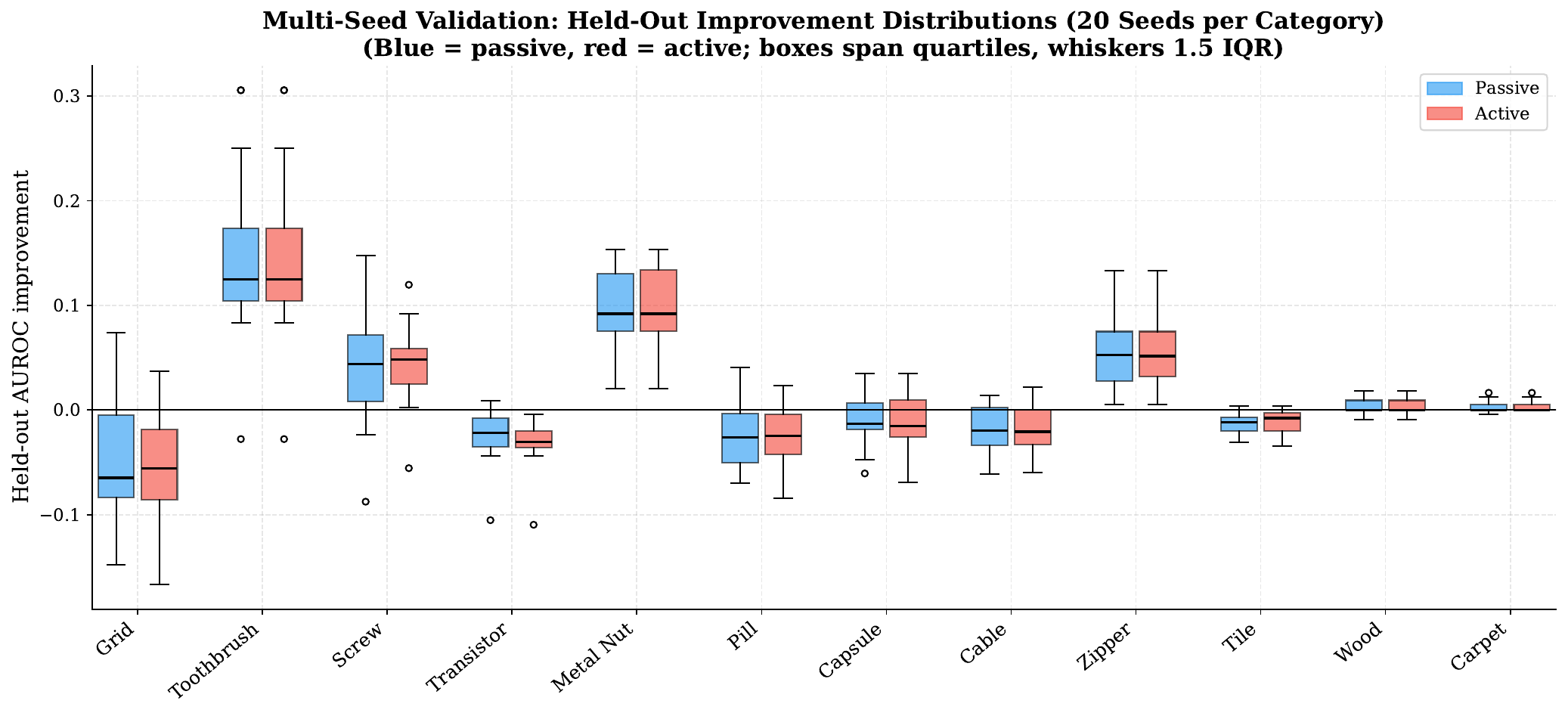}
\caption{Per-seed distributions of held-out AUROC improvement (20 seeds per category; blue = passive, red = active). Toothbrush, metal nut, zipper and screw improve consistently; grid, transistor, pill, capsule, cable and tile centre at or below zero. Passive and active distributions overlap heavily on every category.}
\label{fig:multiseed}
\end{figure}

Every result in this paper is a mean over 20 independent pool/held-out splits (seeds 10--29); multi-seed validation is therefore built into the protocol rather than performed post hoc on selected categories. Figure~\ref{fig:multiseed} shows the full per-seed distribution of held-out improvement for every non-ceiling category and both strategies.

Three observations follow. First, the seed-consistent winners (toothbrush 19/20, metal nut 20/20, zipper 20/20, screw 15/20 seeds improved) are robust well beyond noise. Second, the flat-to-negative categories are consistently flat-to-negative (grid degrades on 15/20 seeds), so their failure is systematic rather than sampling misfortune. Third, on hard categories a single seed can land anywhere within a third of the AUROC scale (toothbrush single-seed improvements range from $-0.03$ to $+0.31$), which explains why earlier single-seed comparisons of querying strategies produced unstable conclusions: individual seeds can and do flip the passive-vs-active winner. Any claim in this domain based on a single split should be treated as anecdote.

\subsection{Label Economics: Is the Gain Just Extra Training Data?}
\label{sec:label_economics}

Since FP corrections append correctly-labelled normal images to the bank, does the held-out gain merely reflect data volume? We test this on toothbrush with matched budgets (20 confirmation seeds): the HITL loop reaches held-out AUROC 0.972 with 12.6 labels on average; labelling the same number of \emph{randomly chosen} pool images reaches only 0.901; appending \emph{all} pool normals (the more-data ceiling) reaches 0.974, but costs $\sim$29 labels, because every pool image must be reviewed to establish which ones are normal. So yes, the gain is delivered through added labelled data, but the loop is what makes that data obtainable. At deployment, unlabelled data cannot be added safely (one mislabelled defect poisons the bank), so the reviewer is the only source; and error-triaged review reaches the more-data ceiling at 43\% of its labelling cost, because random labelling wastes most labels on defects, where corrections are near no-ops. We claim not that the loop beats equivalent training data, but that it is the mechanism by which such data becomes obtainable, cheaply, after deployment.

\subsection{Ablation Study: FP vs.\ FN Correction}
\label{sec:fp_fn_ablation}

To isolate the contributions of FP and FN corrections, we rerun the toothbrush sessions with the reviewer restricted to a single error type (FP-only or FN-only, the skipped type excluded from review entirely) under the full held-out protocol (20 seeds, passive querying). Table~\ref{tab:fp_fn_ablation} reports the result.

\begin{table}[!t]
\caption{FP vs.\ FN correction ablation, toothbrush passive, held-out protocol (means over 20 seeds). Restricted modes exclude the skipped error type from review; $\Delta$ is held-out final minus baseline (0.826).}
\label{tab:fp_fn_ablation}
\centering
\begin{tabular}{lcccc}
\toprule
\textbf{Mode} & \textbf{HO Final} & $\boldsymbol{\Delta}$ & \textbf{HO AUCC} & \textbf{FP/FN corr.} \\
\midrule
FP-only & 0.9722 & $+0.146$ & 0.709 & 7.5/0 \\
Both    & 0.9722 & $+0.146$ & 0.563 & 7.6/5.0 \\
FN-only & 0.8139 & $-0.013$ & 0.004 & 0/8.8 \\
\bottomrule
\end{tabular}
\end{table}

\begin{figure}[!t]
\centering
\includegraphics[width=\columnwidth]{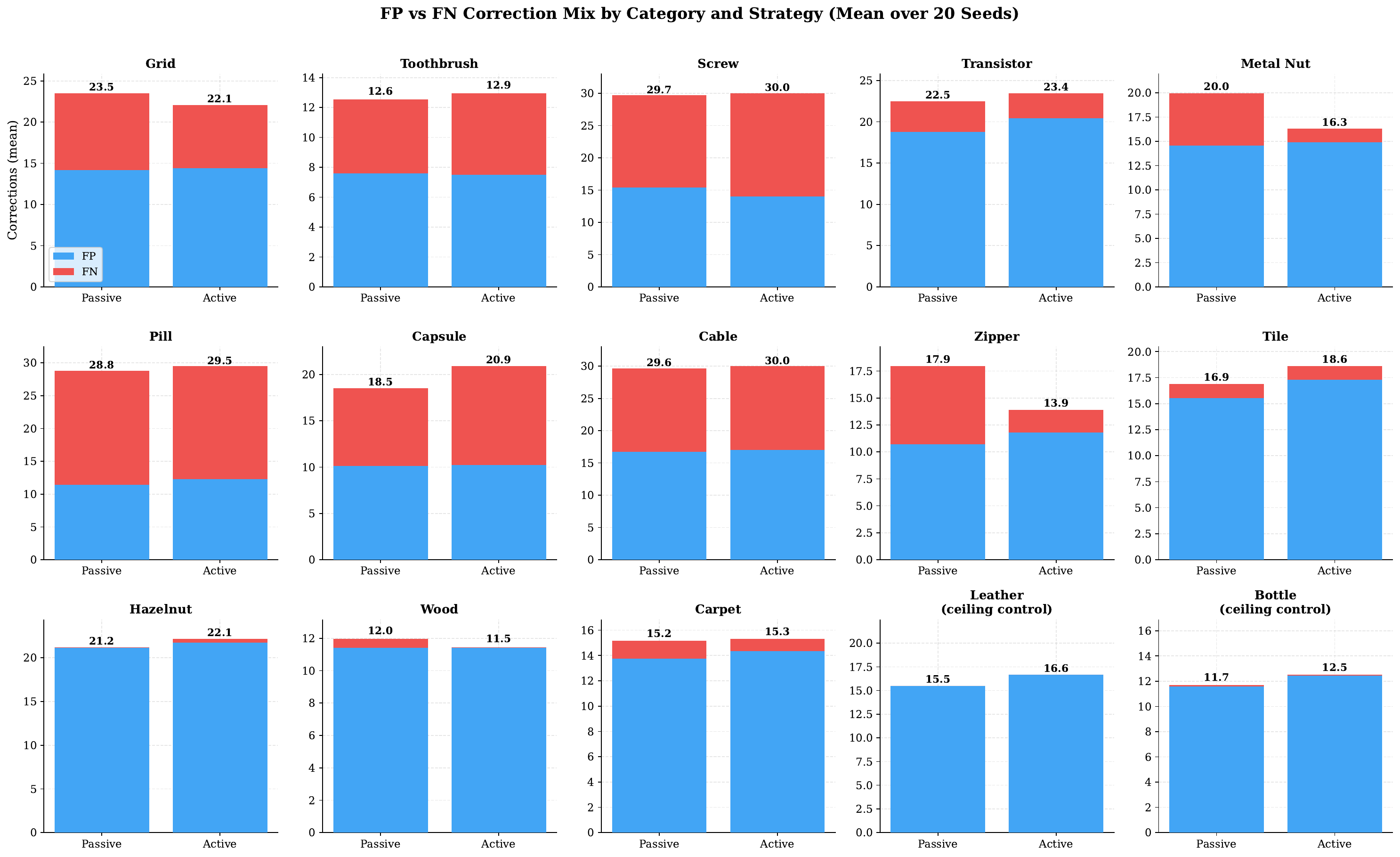}
\caption{FP vs.\ FN correction mix across all 15 categories (mean counts over 20 seeds). Blue = FP, red = FN. Bottle and leather are ceiling controls.}
\label{fig:fp_fn_mix}
\end{figure}

The asymmetry is stark. FP-only correction reproduces the \emph{entire} improvement: its final held-out AUROC equals the combined loop's on every one of the 20 seeds ($+0.146$, Wilcoxon vs.\ zero $p < 10^{-4}$), and it arrives in fewer reviews (7.5 vs.\ 12.6; AUCC 0.709 vs.\ 0.563), because no budget is spent on FN reviews that do not change the destination. FN-only correction achieves nothing: a mean of 8.8 removals per session leaves held-out AUROC at baseline on 13 of 20 seeds, with a residual mean that is slightly \emph{negative} ($-0.013$, unadjusted $p = 0.034$). Removal is, if anything, mildly harmful here. Two causes compound. The first is the bank itself: toothbrush's errors come from gaps in normal coverage, which FP insertion fills directly, whereas FN removal deletes entries from a bank that is already too sparse. The second is a metric mismatch: the removal anchor is selected by 1-NN distance (Eq.~\ref{eq:anchor_patch}) while the image score is the mean-of-nine statistic (Eq.~\ref{eq:anomaly_score}), so the excised neighbourhood is frequently not the one that set the score. Removal is therefore both intrinsically risky on a minimal coreset and often aimed at the wrong location. Figure~\ref{fig:fp_fn_mix} extends the picture to all categories under the multi-seed protocol: every seed-consistent winner has an FP-dominated error pool.

\subsubsection{Why the Anchor Metric Is Not ``Fixed''}
\label{sec:anchor_ablation}

\noindent\textit{Protocol note: like the other parameter ablations, this investigation predates the held-out protocol and was run under pool-only evaluation at a single seed (42).}

Section~\ref{sec:methodology} notes that the removal anchor (Eq.~\ref{eq:anchor_patch}, 1-NN) and the image score (Eq.~\ref{eq:anomaly_score}, mean-of-nine) are different metrics, so removal frequently excises a neighbourhood other than the one that set the score. The obvious remedy is to re-anchor removal on the mean-of-nine patch that actually drives the score. We tested exactly that, on an FN-only correction sequence on toothbrush with $k_{\text{remove}}$ swept over $\{1,2,3,5,8\}$. Re-anchoring does give FN correction a real, non-zero effect, and that effect is destructive: every non-zero removal size collapses AUROC by roughly the same amount, from a baseline of $0.811$ to a final $0.54$--$0.56$, independent both of the anchor metric and of how many points are removed per correction.

The cause is the coreset itself. A false negative scores low precisely because its worst patch sits close to a legitimate normal bank entry; removing that neighbour to fix the one image also pushes away every other normal-looking patch in the same region, including patches from genuinely normal test images. On a bank that is already a minimal 1\% coreset (470 patches for toothbrush) there is no safe removal size, so this is not a magnitude-tuning problem. We therefore retain the 1-NN anchor deliberately. Its mismatch with the scoring metric is what keeps FN correction close to a no-op, and on a minimal coreset a no-op is strictly preferable to a reliable means of destroying the bank. An effective FN mechanism would have to be additive rather than subtractive; the defect-memory pilot (Section~\ref{sec:defect_memory}) tests the most obvious additive design and also fails.

\section{Discussion}
\label{sec:discussion}

\subsection{Why Querying Strategy Barely Matters}

The strategy-equivalence result (Section~\ref{sec:strategy_comparison}) contradicts the standard active learning assumption, and the mechanism explains why. Uncertainty sampling assumes updates are \emph{local}: a boundary-adjacent sample mainly improves the boundary near itself, so sample choice matters. Memory bank corrections are \emph{global}: inserting a false positive's patches changes nearest-neighbour distances for every image, and the review budget exhausts the eligible error pool under any ordering. When every error is eventually corrected and every correction acts globally, order contributes nothing, so uncertainty sampling buys nothing. An earlier single-seed analysis had instead found passive querying \emph{winning} on 9 of 11 categories; multi-seed validation (Section~\ref{sec:multiseed}) exposed that as seed noise, and we retract it as a cautionary example for evaluation practice in this domain.

We also tested a centroid-distance query rule (FP: most centroid-distant normal; FN: most centroid-proximal anomaly) in single-seed pool-protocol experiments predating the held-out protocol: it matched passive on toothbrush, underperformed on metal nut, and was severely counterproductive on cable, in each case through a strong FP-selection bias. Query strategies that measurably beat random ordering for memory bank correction remain an open problem.

\subsection{Negative Result: Defect Memory Bank}
\label{sec:defect_memory}

Could FN corrections contribute symmetrically, by memorising operator-confirmed \emph{defect} patches in a second bank and scoring images by the margin between distance-to-normal and distance-to-defect? A pre-registered pilot (pill, screw, cable; 20 seeds; passive) answers no, decisively: held-out AUROC collapses on pill ($-0.226$) and screw ($-0.299$, worse than the standard loop on 20/20 seeds; both Holm $p<10^{-4}$), with mid-run dips of 0.26--0.49 AUROC, while cable is indistinguishable from the standard loop. A defect bank built from $\sim$28 rejected images over-generalises, pulling large regions of normal appearance toward ``defect.'' Per the pre-registered gate, no full sweep was run. Together with the FP/FN ablation (Section~\ref{sec:fp_fn_ablation}), where FN removal is a measured near no-op, this fixes the paper's central mechanism claim: memory bank correction works by \emph{teaching normality}, not by memorising defects.

\subsection{Bank Growth, and Why the Cap Is Not the Problem}
\label{sec:bank_dynamics}

Bank growth over correction rounds is largest exactly where correction helps most: toothbrush grows $\sim$14$\times$ from its 470-patch original bank, because nearly every inserted patch covers new manifold. For the flat-to-negative group, one hypothesis is the opposite failure: that the cap's random subsampling churns away accumulated corrections. A 20-seed control with the cap removed refutes it: no held-out delta changes except transistor's ($+0.021$). The failures come from the corrections themselves, through redundant insertion into already-covered banks, which is precisely the failure mode the novelty gate removes (Section~\ref{sec:gating_results}).

\subsection{What the Loop Cannot Fix}

Two structural limits survive gating. First, \textbf{globally-defined defects}: cable's \emph{cable\_swap} is locally identical to a normal cable (only the arrangement is wrong), and patch-level correction cannot encode global structure. Second, \textbf{grid}: it inserts 230 patches per correction yet still loses $0.031$ AUROC in the mature-bank setting (Section~\ref{sec:cable_analysis}), so its errors are not a coverage problem; it is the method's one open failure, though it is unharmed at cold start. A general warning follows from the held-out protocol: on every failing category the \emph{pool} metric still rises, so an operator watching in-loop accuracy would believe the loop is working while generalisation worsens. Monitoring must use never-corrected images.

\subsection{Practical Deployment Guidelines}

\textbf{Who:} a quality engineer who already reviews model output. The expert contributes one binary label per reviewed image and needs no ML knowledge, Python, or training infrastructure.

\textbf{When:} enable the novelty gate by default; with it, no MVTec category except grid is significantly harmed, so the loop is safe to leave on. Gains are largest where errors are FP-dominated from an undersampled normal manifold ($+0.05$ to $+0.10$ AUROC gated), and largest of all at cold start, where they reach $+0.25$ (Section~\ref{sec:coldstart_results}). Confirm per category with a held-out check; never gate on in-loop accuracy (see above). Near-ceiling categories need no correction.

\textbf{How:} passive random querying, since active querying adds computation and configuration for no measurable benefit. Effective categories exhaust their errors in 13--20 corrections (screw, with its larger error pool, uses the full 30-round budget), i.e.\ roughly 13--20 minutes of review at one image per minute. Monitor a held-out sample and stop if it trends downward.

\textbf{Infrastructure:} none beyond the deployed model. Bank edits are CPU operations; no training data, no optimisation framework, no ML team.

\section{Conclusion}
\label{sec:conclusion}

We presented a training-free human-in-the-loop framework in which a quality engineer corrects a deployed anomaly detector by reviewing misclassified images: one binary decision per image, applied as a direct memory bank edit. Evaluation across all fifteen MVTec AD categories yields five findings.

\textbf{First, the strongest deployment case is cold start.} From a bank built on ten golden samples, corrections close a median 66\% of the gap to an uncorrected fully trained bank (mean 80\%, raised by three categories that overshoot parity), significantly improving 12 of 15 categories and harming none. Ten samples plus corrections outperform hundreds of samples without them. We are careful not to overstate this: a corrected full-training bank still finishes higher on the same categories (toothbrush 0.927 vs.\ 0.904, metal nut 0.989 vs.\ 0.927), so the result is about what is achievable on day one of a deployment, not about curation dominating data.

\textbf{Second, on mature banks correction is effective where the bank undersamples normal appearance, and only there.} On held-out images the loop never touches, toothbrush improves from AUROC 0.826 to 0.972 in a mean session of roughly 13 single-decision reviews and metal nut from 0.900 to 0.998 in roughly 20, consistently across 20 splits. At one image per minute, that is 13--20 minutes of review. Those endpoints are for the ungated loop; the novelty-gated configuration we recommend for deployment trades some of the gain for safety, reaching 0.927 and 0.989 respectively (third finding below). Five categories sit at ceiling and cannot move, so the mature-bank gains are confined to four of fifteen.

\textbf{Third, honest evaluation requires a held-out protocol, and acting on it produced the novelty gate.} Corrected images enter the bank, so evaluation on reviewed images inflates toward AUROC 1.0 by memorisation. The held-out protocol reveals six categories where the raw loop is flat-to-negative; diagnosing the cause (redundant insertion into already-covered banks) yielded the novelty gate, which eliminates the degradation on five of the six, preserves every winner, and leaves grid as the single open failure.

\textbf{Fourth, querying strategy barely matters.} Passive and active querying are statistically indistinguishable on every category: bank edits act globally, and the budget exhausts the error pool under either ordering. An earlier single-seed finding that passive wins on 9 of 11 categories did not survive multi-seed validation; we retract it explicitly, alongside negative results for centroid-distance querying, for re-anchoring FN removal on the scoring metric, and for a defect-memory second bank. The mechanism is teaching normality, cheaply and in any order.

\textbf{Fifth, the gains equal what the same labels would achieve at training time.} The framework's contribution is not out-performing training data but \emph{producing} correctly-labelled data at deployment, where unlabelled data cannot be added safely, at 43\% of the labelling cost of exhaustive review, and applying it without retraining.

\subsection{Limitations}
\label{sec:limitations}

Our evaluation uses a research implementation on MVTec AD. With per-session feature caching, per-round compute is seconds on hard categories, rising toward a minute as the bank grows; production deployment would use approximate nearest-neighbour indexing. All headline results are means over 20 pool/held-out splits per category (Section~\ref{sec:heldout_protocol}); the centroid-querying investigation and the $k_{\text{remove}}$ parameter ablation predate this protocol and are marked as single-seed, pool-protocol results. Held-out sets are small on some categories ($\sim$13 images for toothbrush), so single-split AUROCs are noisy: read means and seed win-counts (Fig.~\ref{fig:multiseed}), not individual splits. Correction is image-level only; pixel-level correction remains future work.

\subsection{Future Work}
\label{sec:future_work}

\begin{itemize}
\item \textbf{Batched correction:} applying several corrections before rescoring may stabilise complex categories such as cable.
\item \textbf{Production-speed implementation:} approximate nearest-neighbour indexing (FAISS, HNSW) and incremental rescoring would cut per-round latency to seconds, leaving expert review time ($\sim$1 minute per image) as the only bottleneck.
\item \textbf{Pixel-level correction:} letting the expert mark the anomalous region, not just the image label.
\item \textbf{Cross-category transfer:} corrections for one product variant may transfer to related variants.
\item \textbf{Foundation model integration:} applying the correction loop on top of zero-shot detectors \cite{jeong2023winclip,gu2024anomalygpt}.
\item \textbf{Live human trial:} replacing simulated feedback with genuine domain experts (Section~\ref{sec:threats}) is the necessary next validation step.
\end{itemize}

\subsection{Threats to Validity}
\label{sec:threats}

\textbf{Simulated feedback:} the expert is simulated from MVTec ground-truth labels and is always right. Real reviewers disagree with ground truth in ambiguous cases, fatigue, and drift; our results are an upper bound under perfect feedback, and a live trial is required before industrial claims.

\textbf{Operator error is costliest at cold start:} relaxing the perfect-feedback assumption does not cost the same in every regime. A single ``normal'' label on a true defect inserts up to 784 defect patches into the normal bank. Against a mature bank near the 12{,}000-patch cap that is a few per cent contamination; against a 78-patch cold-start bank it is an order of magnitude more feature mass than the bank itself. Our strongest result is therefore also the one most exposed to imperfect feedback, and a live trial should establish an operator error rate, and the value of an escalation option that applies no correction, before the small-bank regime is deployed.

\textbf{Single dataset:} MVTec AD spans 15 diverse categories but not the full range of factory conditions (lighting variation, camera drift, novel defect types). Findings should be revalidated on proprietary data before being treated as general.

\textbf{Mechanism specificity:} direct bank editing applies to memory-bank detectors (PatchCore and coreset-based relatives). Reconstruction-, flow-, or classifier-based detectors expose no equivalent editable structure; we claim nothing beyond the memory-bank family.

\textbf{Timing:} reported wall-clock times are dominated by research-grade GPU rescoring. ``Minutes of review'' describes the human component and assumes the production-speed implementation above.

Prior work has modified PatchCore banks automatically; what this paper establishes is different in kind: the person who already reviews the errors can now fix them immediately, without an ML team. Where a formal retraining cycle takes weeks to months, a correction session takes 13--20 minutes. That shift, in both who can correct the model and how fast, is the framework's most useful property.

\section*{Data and Code Availability}
All experiments use the publicly available MVTec AD benchmark \cite{bergmann2019mvtec}, obtainable from the MVTec Software GmbH website under its research licence; no proprietary or personal data were used. The correction-framework implementation and the experiment scripts that reproduce every reported result are available from the corresponding author on reasonable request.

\section*{Acknowledgment}
This research received no external funding.

\bibliographystyle{IEEEtran}
\bibliography{references}

\end{document}